\documentclass[10pt,twocolumn,letterpaper]{article}

\usepackage{iccv}
\usepackage{times}
\usepackage{epsfig}
\usepackage{graphicx}
\usepackage{amsmath}
\usepackage{amssymb}
\usepackage{array}
\usepackage{mathtools}
\usepackage[export]{adjustbox}
\usepackage{tabulary}
\usepackage{caption} 
\usepackage{xcolor}
\usepackage[font=footnotesize,labelfont=bf]{caption}

\usepackage[pagebackref=true,breaklinks=true,letterpaper=true,colorlinks,bookmarks=false]{hyperref}

\iccvfinalcopy 

\ificcvfinal\fi
\begin{document}

\title{A Distraction Score for Watermarks}

\author{Aurelia Guy \\
Google \\
{\tt\small liaguy@google.com}
\and
Sema Berkiten \\
Google \\
{\tt\small berkiten@google.com}
}

\maketitle

\begin{abstract}
  In this work we propose a novel technique to quantify how distracting watermarks are on an image. We begin with watermark detection using a two-tower CNN model composed of a binary classification task and a semantic segmentation prediction. With this model, we demonstrate significant improvement in image precision while maintaining per-pixel accuracy, especially for our real-world dataset with sparse positive examples. We fit a nonlinear function to represent detected watermarks by a single score correlated with human perception based on their size, location, and visual obstructiveness. Finally, we validate our method in an image ranking setup, which is the main application of our watermark scoring algorithm.
  
\end{abstract}

\section{Introduction}

Watermarking images is a common practice for both businesses and individuals to mark
ownership.  Oftentimes, the watermark is small and can be ignored, only taking up a
small portion of the image and not taking away from the overall image quality.
However there is a subset of watermarks that greatly impact visual quality. For these images,
the visual quality is greatly decreased by their watermarks. In this work we
aim to infer an image score that encodes its watermarks' perceptual impact. One of the major motivation for such a score is to demote images based on how distracting their watermarks are for an image ranking task such as for image ranking in Google Maps. Our objective is to score images with high precision for the broad watermark class, which includes watermarks that range from mostly transparent to fully opaque, as well as watermarks of arbitrary \textit{shapes}, \textit{texts}, and \textit{patterns}. 

Our work is performed on a proprietary dataset of 200k annotated images,
taken by individuals or businesses. As the majority of images in our full database do not contain
watermarks, we additionally consider the performance of our model in terms
of real world data, which we estimate to contain less than 10\% watermarked images by randomly sampling our database of billions of images.

\paragraph{DeepLab.} In this work, we formulate our problem of detecting watermarks as a combination of binary image classification, used to improve our precision, and a semantic image segmentation task. To train our model, we use a variation of the DeepLab architecture~\cite{DeepLab}. The DeepLab framework implements
atrous convolution with an encoder-decoder structure~\cite{DBLP:journals/corr/ChenPSA17, DBLP:journals/corr/abs-1802-02611} to segment objects
at multiple scales~\cite{DBLP:journals/corr/ChenPKMY14, DeepLab}. We utilize a variation of this model with the final Fully Connected CRF
layer replaced by likelihood thresholding. The class likelihood scores for all scales are merged and
thresholded to obtain the segmentation labels.

\begin{figure*}[ht!]
\textbf{Hybrid Model}\par\medskip
\vspace{-2pt}
\resizebox{\linewidth}{!} {
\centering
\vspace{-12pt}
\includegraphics[scale=0.3,frame]{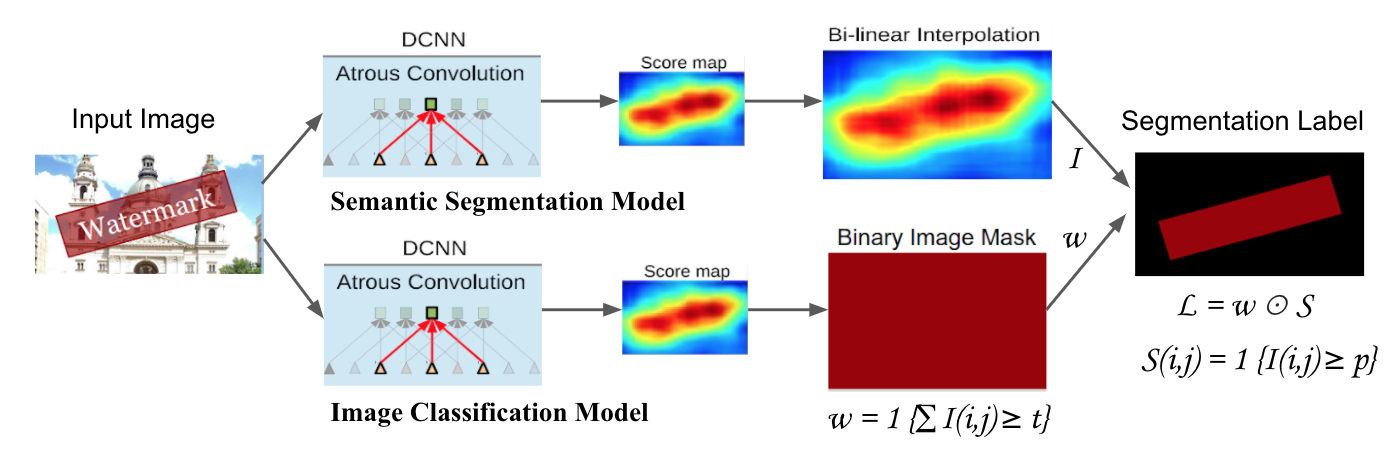}}
\caption{Diagram of the proposed hybrid model. A watermarked image is input to the image segmentation and classification models based on the DeepLab architecture~\cite{DeepLab}. binary image mask ($w$) is a black or white image based on the number of pixels is below or above ($t$). The final segmentation label ($L$) is obtained by taking inner product of the binary image mask $w$ and segmentation map $S$ where $S$ is a binary image  obtained by thresholding ($p$) score map $I$.}
\label{fig:HybridModel}
\vspace{-10pt}
\end{figure*}

In this work, our contribution is twofold:
\begin{itemize}
    \item{A two-tower CNN architecture to train a segmentation model detecting watermarks with high reliability as shown in Figure~\ref{fig:HybridModel};}
    \item{A scoring function to convert detected watermarks into a single score which represents human response to the watermarks' obstructiveness.}
\end{itemize}


\section{Related Work}
\paragraph{Text Detector.}
An alternative approach for watermark detection is a text detector,
as many watermarks contain text. In section~\ref{sssec:textdetectors}, we compare the inference results for Google Cloud API's text detector~\cite{GoogleCloudAPI} to our trained DeepLab-based model.
We demonstrate that the frequency of false positives in the results from a
text detector model makes it unsuitable for our watermark detection task.

\paragraph{Watermark Removal.} Another interesting and related problem to our task is watermark removal to recover the original image. Dekel et al.~\cite{WatermarkRemoval}
demonstrate this approach with transparent watermarks of limited, predetermined types. This work leverages the consistent manner through which watermarks are added to images, as similar versions of logos are added to millions of images on the web. The removal problem is treated as a multi-image matting problem~\cite{ImageMatting, ImageVideoMatting} to separate the images into foreground and background, where the image structures of the common watermarks in a collection of images are utilized to detect and remove the watermarked regions. This work generalizes to positional variations and subtle geometric and color variations only. Alternative watermark removal techniques include the use of image inpainting ~\cite{inpaiting, attackingWatermark} and Independent Component Analysis ~\cite{ICA} to separate the watermark from the image.  Braudaway et al. ~\cite{VisibleImageWatermark} introduced a digital watermarking technique for applying a watermark over an image while not obstructing any of the image detail. Similarly, Kankanhalli et al.~\cite{AdaptiveVisibleWatermarking} developed a method to construct pleasant and unobtrusive watermarked images based on the content of image. Watermark removal has also been demonstrated by Belmont~\cite{WatermarkAdditiveRemoval} who adds a mask to an image and trains the network to find where the mask was overlaid. This architecture does not generalize well to watermarks that differ significantly from the generated watermark, however, and requires that the watermark is overlaid in an additive fashion. Related to removing watermarks in images is watermark removal in videos, which relies on temporal consistency~\cite{VideoLogoRemoval, AutomaticTvLogoDetection,  AutomaticVideoLogoDetectionAndRemoval}. In our work, we investigate detecting watermarks of varying image structures and do not restrict our model to synthetic or repeating data.

\paragraph{Mask R-CNN.}
Similar approaches to our image segmentation task are Mask R-CNN and DeepMask, which construct masks for image patches used for object proposals~\cite{DBLP:journals/corr/HeGDG17, DBLP:journals/corr/PinheiroCD15}. In addition, Pinheiro et al.~\cite{DBLP:journals/corr/PinheiroLCD16} introduces a feed-forward architecture which outputs a coarse mask encoding in the feed-forward pass. Dai et al.~\cite{DBLP:journals/corr/DaiHS15} uses a cascading architecture with three networks, including a mask estimation. However, the networks introduced in these works share convolutional features during training. We train our mask network separately on a dataset with negative examples and prioritize per pixel precision over accuracy. Our mask and segmentation networks are then combined after training. As a result, our proposed network improves ranking accuracy for data with few positive examples. 

\begin{figure*}[ht!]
\textbf{Watermark Semantic Segmentation Models}\par\medskip
\resizebox{\linewidth}{!} {
\includegraphics[scale=0.22]{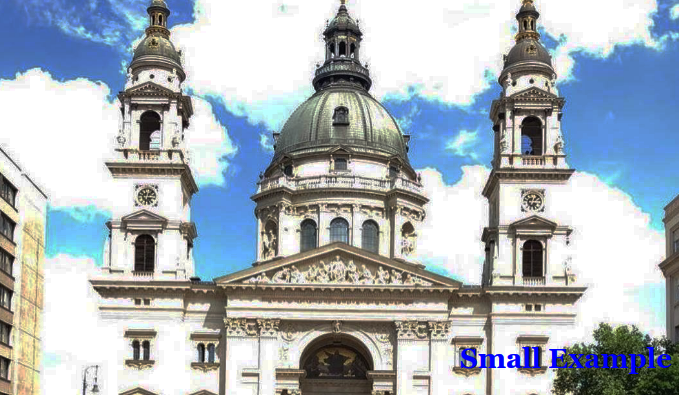}
\includegraphics[scale=0.22]{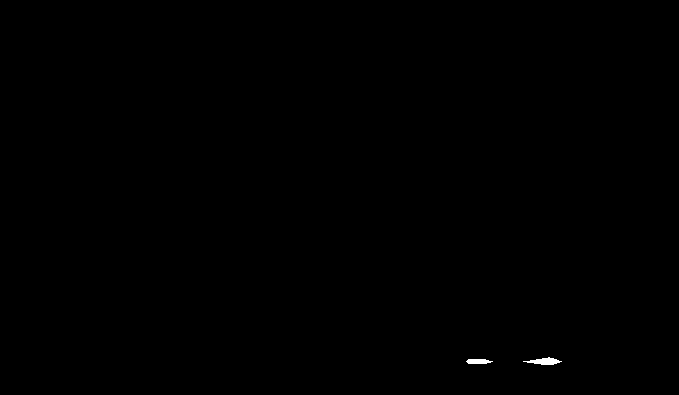}
\includegraphics[scale=0.22]{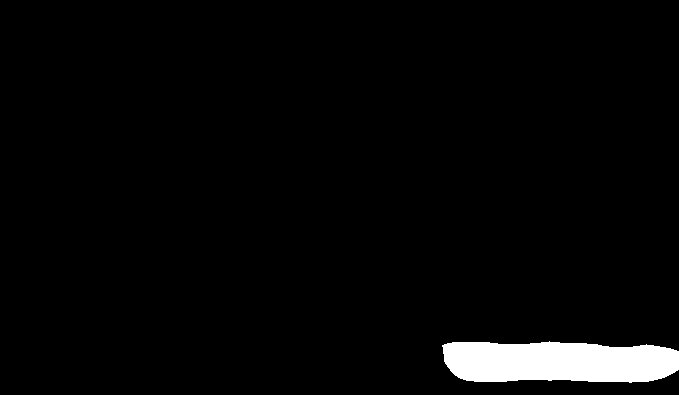}}
\vspace{-8pt}
\caption{Semantic segmentation predictions for a small watermark in the corner of the image (\textit{left}). This image
demonstrates the hybrid model's higher watermark pixel recall. Our strict image classification model's low pixel recall results in the segmentation label (\textit{middle}) with significantly fewer predicted watermark pixels than our hybrid model's predicted segmentation label (\textit{right}).}
\label{fig:PixelRecallComparison}
\vspace{-10pt}
\end{figure*}

\section{Method}

In the initial training of our DeepLab model for image segmentation, we observed
a high frequency of image false positives. Images that contained signs,
billboards, menus, and other text or logos that resembled watermarks were
falsely predicted to contain watermarks. In order to reduce the frequency
of these predictions, we developed a model that favored precision over
recall, increasing the likelihood threshold for a watermark pixel and adding
negative examples to our training dataset. 

A limitation of our strict model is a decrease in per pixel recall for images that contained
watermarks. We observe that by penalizing watermark pixel loss for images that
only contained 6\% watermark pixels, we are biasing our model towards negative predictions. DeepLab's semantic
segmentation model calculates the total loss at each iteration as the sum of the
loss of each class. Since the non-watermark class comprises 97.5\% of the total
pixels during training, watermark false negatives are highly favored. This results in image segmentation outputs that are very conservative in their prediction of watermark pixels,
only predicting a small percentage as watermarks as shown in Figure~\ref{fig:PixelRecallComparison}. 

Although the prediction generated by DeepLab is sufficient to identify
the existence of a watermark in the image, it
significantly reduces the accuracy of our watermark scoring step. In order to increase the accuracy, we propose a hybrid model as described in Section~\ref{sssec:HybridModel}.
For watermark scoring described in Section~\ref{sssec:score}, we consider the size and location of the watermark, which
requires an approximate segmentation of the image into watermark and non-watermark pixels.

\subsection{Semantic Image Segmentation} \label{sssec:SemanticModel}
In order to generate a distraction score for watermarks, we must first detect watermark pixels. To achieve this, we train a model on images that contained watermarks.~As a pre-processing step, images are scaled to be of uniform size. 
By eliminating negative examples and
fixing the watermark class frequency, we reformulate the watermark scoring problem to a
semantic segmentation one. This model is proficient in its segmentation of
images into watermark and non-watermark pixels, allowing for accurate watermark
scoring of images containing watermarks. However, as this model is trained
on positive examples, this model alone has a higher false positive rate and lower image precision than our hybrid model, see Section~\ref{sssec:metrics} for details.

\subsection{Image Classification} \label{sssec:BinaryModel}
To address the high rate of false positives, we train a second, separate image classification model. This model is trained on a dataset of images that both do and do not contain watermarks and is meant to discriminate between the two. We modify the DeepLab architecture~\cite{DeepLab} by replacing the final Fully Connected CRF layer by likelihood thresholding, And we use a threshold of 0.75 to classify a pixel as containing a watermark.
The resulting
segmentation prediction is transformed into a binary mask by thresholding
the number of predicted watermark pixels in the image. We set the threshold
at a small fixed value based on the image size and observed noise in the
segmentation label map.


\subsection {Hybrid Model} \label{sssec:HybridModel}
Finally, we combine the two models described in Sections~\ref{sssec:SemanticModel} and~\ref{sssec:BinaryModel} by masking the predicted segmentation label from the image 
segmentation model with the binary image classification.
This is equivalent to multiplying the pixel segmentation labels for the image $I$ by $w_I$,
where $w_I$ is binary classification $\{0,1\}$ for the image $I$. 
\begin{equation*}
\begin{aligned}
&L(i,j) = w_I \cdot S(i,j)
\end{aligned}
\end{equation*}
where $S(i,j)$ is the semantic segmentation prediction of image $I$ at pixel $(i,j)$ and $L(i,j)$ is the final segmentation label of the image for our hybrid model.

We designed this model to improve the accuracy of image comparison for images from a dataset with sparse positive examples. For each image, a final watermark distraction score is calculated as detailed in Sec.~\ref{sssec:score}. We then define true positives for this image comparison as:

\begin{equation*}
\begin{aligned}
\sum_{y_i > y_j}
\begin{cases}
  1 \quad\text{if } \text{Score}(I_i) > \text{Score}(I_j)\\
  0 \quad\text{otherwise}
\end{cases}
\end{aligned}
\end{equation*}
where for a pair of images, $y_i$ is the true score of image $I_i$ and $y_j$ is the true score of image $I_j$. We seek to maximize the number of image pairs that we score correctly. That is, predicting a higher score for the image with more visually distracting watermarks. For a scoring function where we compute the normalized area of watermarked pixels, this is equivalent to maximizing the Intersection over Union (IOU), as defined in Section~\ref{sssec:metrics}, of images that contain watermarks, while minimizing the number of incorrectly classified images. This reformulation of our problem motivates our hybrid model design, since for a segmentation label $L_i$ of image $I_i$:
\begin{equation*}
\begin{aligned}
 \max IOU(L_i), \; \; I_i \in Positive Examples 
\end{aligned}
\end{equation*}
 will be maximized for a model with the highest pixel IOU value on images containing watermarks. In the design of our semantic segmentation model, the pixel IOU was maximized. In addition, minimizing the number of image False Positives ($iFP$) and image False Negatives ($iFN$), as defined in Section~\ref{sssec:metrics}, of our model is equivalent to maximizing:

\begin{align*}
\frac{iTP}{iTP + iFP + iFN}, \; \; I \in Balanced Dataset
\end{align*}
where $BalancedDataset$ is a dataset containing both watermarked and non-watermarked images. For a dataset where $PositiveExamples \ll NegativeExamples$, we can assume $iFN \ll iFP$. Therefore, we can approximate this expression with:
\begin{align*}
\frac{iTP}{iTP + iFP}, \; \; I \in SparseDataset
\end{align*}

This expression is the image precision of our dataset, which our image classification model maximizes. By design, our hybrid model obtains a higher precision than the image segmentation model, as its $iFP$ is upper bound by the $iFP$ of the image classification model. In addition, for images that are not
$iFN$ of the image classification model, the pixel recall is determined by the semantic
segmentation model, which has a high pixel recall, as shown for an example in Figure~\ref{fig:PixelRecallComparison}. As a result, by combining the image classification and semantic segmentation models, we maximize the accuracy of our scoring function in the context of image comparison. A diagram of our proposed
hybrid model is shown in Figure~\ref{fig:HybridModel}.

\subsection{Watermark Distraction Scoring} \label{sssec:score}
Finally, we score images based on the total area and locations of their detected watermarks to represent their perceptual impact with a single score.
This scoring function maps segmentation labels into a score in the range of $[0,1]$, where images with a score of $0$ do not contain
watermarks and images with a score of $1$ contain watermarks that are very distracting
and visually obstructive.
We characterize the weight of the watermark's location as an isotropic 2D Gaussian function centered at the center of the image. We incorporate the total
area of image $I$'s watermarks into our image score by computing the weighted sum of
the segmentation labels $L(i,j)$ with the Gaussian weights $g(i,j)$ as follows:
\begin{equation*}
\begin{aligned}
& G(\sigma, L) = \sum_{i,j \in L} g(i,j) L(i,j)
\end{aligned}
\end{equation*}
We also assume that users' scoring of image quality is not necessarily linear with respect to this weighted sum.
We therefore fit the output to the user responses with a sigmoid function as:
\begin{equation*}
\begin{aligned}
& Score(I) = \frac{1}{1 + e^{- \lambda (G(\sigma, L) - \alpha)}}
\end{aligned}
\end{equation*}
where $\alpha$ is the bias, $\sigma$ is the standard deviation of the Gaussian
and $\lambda$ is the steepness of the sigmoid function. $G(\sigma, L)$ is
the weighted sum of elements $(i,j)$ in $L$ with the value of the Gaussian at $(i,j)$.
We optimize for $\alpha$, $\sigma$, and $\lambda$ by minimizing a Mean Squared Error: 
\begin{equation*}
\begin{aligned}
& \underset{\lambda, \sigma, \alpha}{\text{argmin}}
& & MSE(\frac{1}{1 + e^{- \lambda (G(\sigma, L) - \alpha)}}, y)
\end{aligned}
\end{equation*}
where $y$ is the watermark score obtained from user responses. To fit our scoring function, we obtained a ground-truth dataset of 10k segmentation labels with
corresponding watermark scores. Scores were obtained by displaying watermarked images and
asking reviewers to give them a discrete score between $[0,3]$ based on the size, locations, and obstructiveness of their watermarks with $0$ being no watermark and $3$ being a large and very obtrusive watermark.

\section{Experiments}
\subsection {Datasets}
\label{sssec:dataset}
We collected a proprietary dataset of 200k annotated images. Of these images, 62.5\% contain watermarks. For images containing
watermarks, the watermark pixel frequency is 6.011\%. We randomly split our labelled data into three sets: training (80\%), evaluation (10\%), and validation (10\%). We train our model on 160k annotated images. During training, we evaluate our model
on an evaluation set of 20k images. And after training, we evaluate our model on a validation set of 20k images. 

\begin{figure}
\centering
\includegraphics[width=0.43\textwidth]{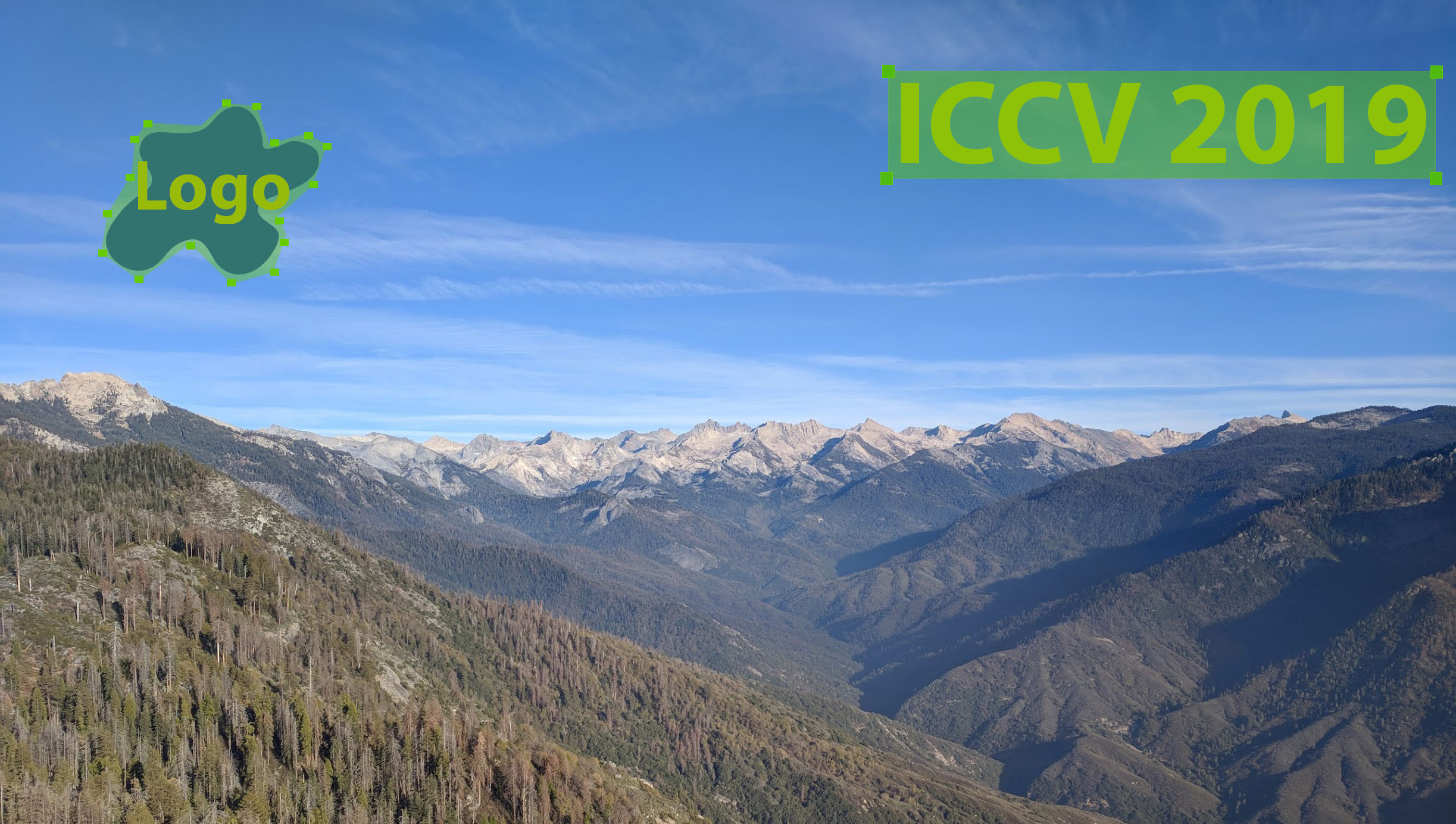}
\vspace{-4pt}
\caption{Example of an annotated image used as a segmentation label. Watermarks on the top right and left corners of the image are annotated with two polygons with an arbitrary numbers of vertices.}
\vspace{-12pt}
\label{fig:AnnotatedImage}
\end{figure}

The images in our dataset are annotated as shown in Figure~\ref{fig:AnnotatedImage}. Annotators
are asked to draw polygons around each watermark region in the image, which are then converted into
a binary segmentation label. If an image contains no watermark annotators are instructed to add no labels and we interpret the result accordingly.

\subsection{Evaluation Metrics} \label{sssec:metrics}
We consider both pixel level and image level metrics when evaluating our model. 
\paragraph{Pixel Metrics.} We define pixel precision, recall and IOU as:
\begin{equation*}
\begin{aligned}
Precision &= \frac{TP}{TP + FP} \\
Recall &= \frac{TP}{TP + FN} \\
IOU &= \frac{TP}{TP + FP + FN}
\end{aligned}
\end{equation*} 
where a True Positive ($TP$) is a pixel that is correctly classified as a watermark pixel and  a False Positive ($FP$) is a pixel that is incorrectly classified as a watermark pixel.
\paragraph{Image Metrics.} We define image precision and recall as:
\begin{equation*}
\begin{aligned}
iPrecision = \frac{iTP}{iTP + iFP} \\
iRecall = \frac{iTP}{iTP + iFN}
\end{aligned}
\end{equation*}
where an image True Positive ($iTP$) is an image that is correctly predicted to contain a watermark.

\paragraph{Estimated Image Precision.}
In this paper, we assume that watermarked images comprise approximately 10\% of our entire database based on a small subset of our entire database.
Although our hybrid model's image recall is lower than the segmentation model, it provides significant improvement in $iPrecision$ for
datasets with sparse positive examples.
We compute the estimated image precision ($ePrecision$) of our model for datasets with different percentages of positive examples as:
\begin{equation*}
\begin{aligned}
& ePrecision = \frac{\beta iTP}{\beta iTP + (1-\beta) iFP}
\end{aligned}
\end{equation*}
where $\beta$ is the true fraction of watermarked images in our dataset, and $iTP$ and $iFP$ are computed from a balanced dataset of $50\%$ watermarked images.
If we compute $ePrecision$ with $\beta=0.1$ for the semantic segmentation model shown in Table~\ref{table:CompareImageModels}, only 26.13\% of detected images would actually contain watermarks.
On the other hand for the strict hybrid model, 79.51\% of detected images would contain watermarks.
This improvement increases the $ePrecision$ of our model, as the majority
of images with a non-zero watermark scores are true positives. We compare $ePrecision$ values for our hybrid model to the image segmentation model in Section~\ref{sssec:balanced}.

\begin{table}
\centering
\begin{tabular}{|c|c|}
\hline
Dataset & mIOU \\
\hline\hline
Watermarked (ours) &  76.16 \\
Balanced Watermarked (ours) & 63.04 \\
\hline
PASCAL VOC 2012 val & 77.69 \\
PASCAL-Context dataset & 45.7 \\
PASCAL-Person-Part & 63.10 \\
Cityscapes dataset & 70.4 \\
\hline
\end{tabular}
\caption{Comparison of the mean IOU for our watermark datasets to the datasets in Chen et al.~\cite{DeepLab} for DeepLab model.}
\vspace{-10pt}
\label{table:MeanIouTable}
\end{table}

\paragraph{Mean IOU Comparison.}
For our image segmentation model, we obtain a mean IOU of 76\% on our validation, similar to~\cite{DeepLab},
as shown in Table~\ref{table:MeanIouTable}. This mean IOU metric demonstrates the accuracy
and precision of our watermark scoring function. For our image classification model, our mean IOU
is lower because we trade pixel recall for image precision. If we weight our mean IOU
to balance the frequency of the watermark and non-watermark classes, where watermark pixels make up 6\%
of image pixels (as detailed in Sec.~\ref{sssec:dataset}), our mean IOU is 63.04\%. 

The lower value demonstrates the difficulty
of the watermark problem, as watermarks vary significantly in shape, pattern and transparency. Moreover we would also like to note that mean IOU is not very high because of the fact that most of the annotated images have background pixels included in the labels such as in Figure~\ref{fig:AnnotatedImage}.

\subsection{Semantic Segmentation Evaluation}

We train our semantic segmentation model on 160k watermarked images, as detailed in Sec.~\ref{sssec:dataset}. The results are shown in Table~\ref{table:CompareModels}. This model attains a high pixel precision and recall, with
an IOU of $65$, comparable to the mean IOU values reported in~\cite{DeepLab}.
However, when we add negative examples to the validation set, the pixel precision
decreases to 69\%, with an image precision of only 76\%. For images with watermarks,
this model proficiently segments the images into watermark and non-watermark regions.
However, this model is unable to effectively filter non-watermarked images, as shown in Table~\ref{table:CompareModelsWithNegatives}.
Our image classifier maintains a 95\% pixel precision after we
include negative examples in our validation set. In order to obtain this high precision,
its watermark recall is only 43\%. For our hybrid model, since the image false positives
are upper bound by the image false positives of our image classifier, it maintains
an image precision of 97\%, as shown in Table~\ref{table:CompareImageModels}. In addition, it attains a pixel recall of 70\% and
pixel precision of 83\%.

\begin{table}
\resizebox{\columnwidth}{!} {
\begin{tabular}{|c|c|c|c|}
\hline
Model & Precision & Recall & IOU \\
\hline\hline
Semantic Segmentation & 82.05 & 75.68 & 64.93 \\
Image Classifier & 95.42 & 43.04 & 42.17 \\
\hline
Hybrid Classifier & 83.29 & 69.60 & 61.07 \\
\hline
\end{tabular}}
\caption{Pixel precision, recall and IOU for dataset with only watermarked images.}
\vspace{-8pt}
\label{table:CompareModels}
\end{table}

\begin{table}
\resizebox{\columnwidth}{!} {
\begin{tabular}{|c|c|c|c|}
\hline
Model & Precision & Recall & IOU \\
\hline\hline
Semantic Segmentation & 68.97 & 75.68 & 56.46 \\
Image Classifier & 94.58 & 43.04 & 42.00 \\
\hline
Hybrid Classifier & 81.10 & 69.60 & 59.80 \\
\hline
\end{tabular}}
\caption{Pixel precision, recall and IOU for dataset with 62.5\% watermarked images. Our hybrid classifier has improvement over semantic segmentation model in pixel precision.}
\vspace{-8pt}
\label{table:CompareModelsWithNegatives}
\end{table}

\begin{table}
\resizebox{\columnwidth}{!} {
\begin{tabular}{|c|c|c|c|c|c|}
\hline
Model & iPrecision & iRecall & ePrecision\\
\hline\hline
Semantic Segmentation & 76.10 & 97.20 & 26.13 \\
Image Classifier & 97.16 & 84.63 & 79.51 \\
\hline
Hybrid Classifier & 97.21 & 84.61 & 79.51 \\
\hline
\end{tabular}}
\caption{Image precision and recall for a dataset with 62.5\% watermarked images, weighted by image class frequency. We show our estimated image Precision (ePrecision) for a dataset with 10\% watermarked images as well. Our hybrid classifier maintains the precision of our image classifier.}
\label{table:CompareImageModels}
\vspace{-8pt}
\end{table}

\subsection{Semantic Segmentation with a Balanced Set} \label{sssec:balanced}
As an alternative approach to our hybrid model, we train our semantic segmentation model
with negative examples to obtain similar per pixel metrics to our hybrid model.
A comparison of these metrics are shown in Table~\ref{table:ComparisonBalancedPixel}.
For our hybrid classifier, we match per pixel metrics obtained with semantic segmentation.
In a balanced validation set, the image precisions of these two models are comparable. Our semantic segmentation model obtains 94.24\% image
precision with a balanced validation set, only a 3\% decrease in precision from our
hybrid model with 97.21\% precision.

However, when we consider our true dataset, our hybrid model significantly outperforms the semantic
segmentation model. The semantic segmentation model has over double the false positives as our hybrid model, with 471 $iFP$ as opposed to only $201$ iFP in our hybrid model.
In a dataset with sparse positive examples, this increase in false positives has a large effect on the image precision.
For our true dataset with 10\% watermarked images, only 65\% of detected images would contain watermarks.
The estimated image precision of our hybrid model is higher, with 80\% of detected images containing watermarks, as shown in Table~\ref{table:ComparisonProjectedPrecision}.

\begin{table}
\resizebox{\columnwidth}{!} {
\begin{tabular}{|c|c|c|c|}
\hline
Model & Precision & Recall & IOU \\
\hline \hline
Semantic Segmentation & 85.62 & 67.99 & 61.03 \\
Hybrid Classifier & 83.29 & 69.60 & 61.07 \\
\hline
\end{tabular}}
\caption{Pixel precision, recall and IOU for watermarked images.}
\label{table:ComparisonBalancedPixel}
\vspace{-8pt}
\end{table}

\begin{table}
\resizebox{\columnwidth}{!} {
\begin{tabular}{|c|c|c|c|c|c|}
\hline
Model & iPrecision & iFP & ePrecision \\
\hline\hline
Semantic Segmentation & 94.24 & 471 & 64.51 \\
Hybrid Classifier & 97.21 & 201 & 79.51 \\
\hline
\end{tabular}}
\caption{Image precision and number of image False Positives with 62.5\% watermarked images, and estimated image Precision with 10\% watermarked images.}
\label{table:ComparisonProjectedPrecision}
\vspace{-8pt}
\end{table}

\subsection{Text Detectors} \label{sssec:textdetectors}
\begin{figure}
\resizebox{\linewidth}{!} {
\includegraphics[scale=0.22]{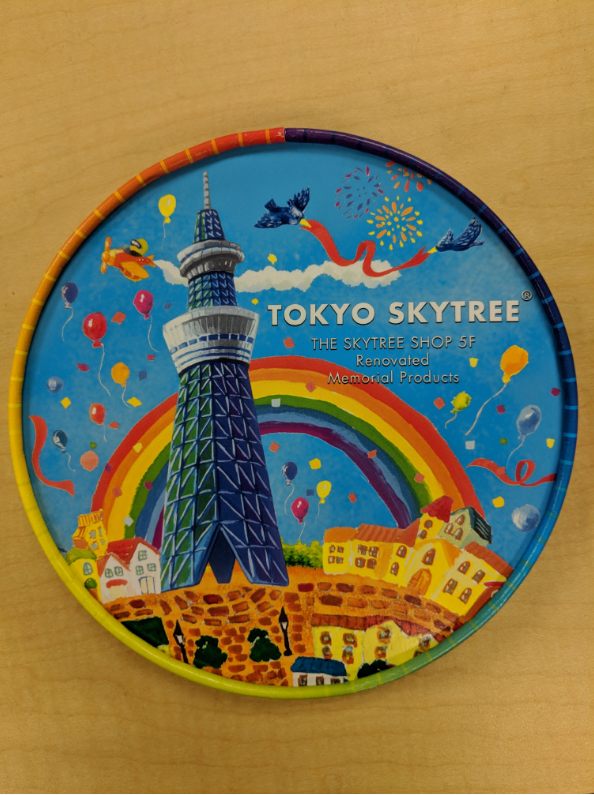}
\includegraphics[scale=0.22]{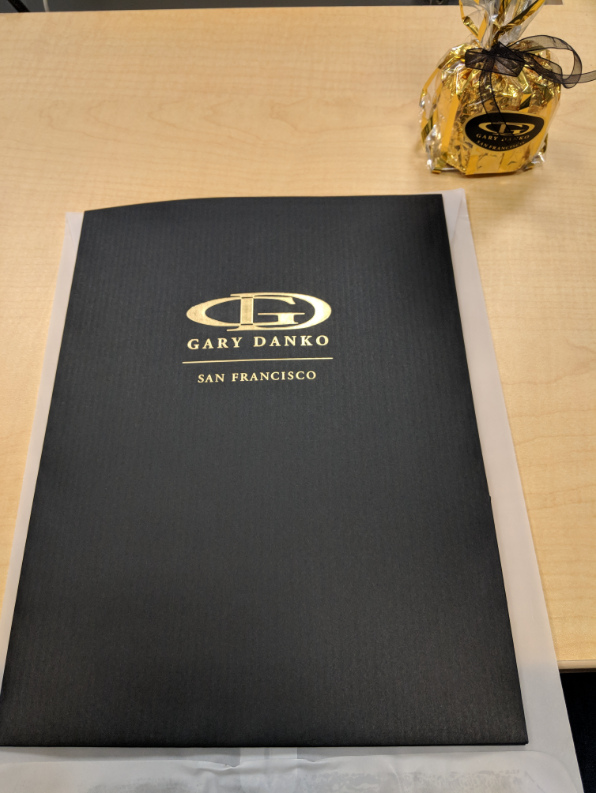}
\includegraphics[scale=0.22]{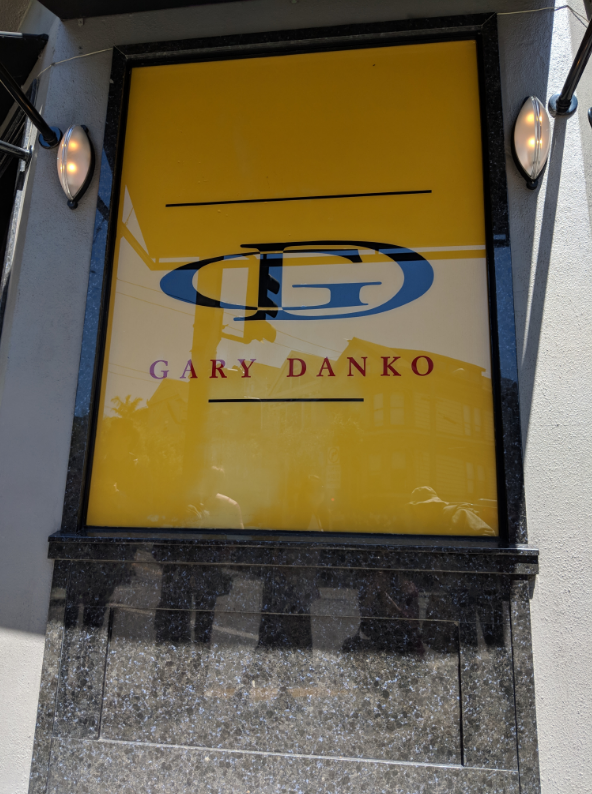}}
\caption{Google Cloud API's text detector~\cite{GoogleCloudAPI} falsely predicts images to contain watermarks. Our hybrid model predicts these images to not contain watermarks.}
\label{fig:TextComparison}
\vspace{-8pt}
\end{figure}

We evaluate our validation set on Google Cloud API's text detector~\cite{GoogleCloudAPI}. This text detector is an OCR (optical character recognition)
system for extracting text from images. We consider a text detector as an
alternative for detecting watermarks, as they commonly contain text. However,
we identify two limitations to using a text detector for general watermark
detection. Firstly, text in watermarks only comprises a subset of watermark
pixels, resulting in a low per pixel recall as shown in Table~\ref{table:TextDetectorTable}.
Secondly, non-watermark text in images increases the false positive rate. We observe
a high false positive rate for non-watermark text on billboards, menus, signs, and fliers.
Figure~\ref{fig:TextComparison} shows examples of the text detector's false positives 
that our hybrid model correctly predicts to not contain watermarks.

\begin{table}
\resizebox{\columnwidth}{!} {
\begin{tabular}{|c|c|c|c|c|c|}
\hline
Model & Precision & Recall & IOU & iPrecision & iRecall \\
\hline\hline
Text Detector & 33.08 & 15.35 & 11.71 & 65.91 & 61.89 \\
Hybrid & 81.10 & 69.60 & 59.80 & 97.21 & 84.61 \\
\hline
\end{tabular}}
\caption{Comparison of precision and recall metrics for text and hybrid classifers.}
\label{table:TextDetectorTable}
\vspace{-8pt}
\end{table}

\subsection{Results on Images From a Different Domain}
We also run our algorithm on publicly available images containing watermarks on Flickr~\cite{Flickr}. Table~\ref{table:FlickrTable} shows the original images, detected watermarks, and their final watermark scores. The results show the ability of our model to detect various kinds of watermarks including text (even when rotated), logos, and transparent and solid watermarks. Also, the estimated watermark scores are highly correlated with the watermarks' perceptual impact on the images, such as when the watermark is covering most of the image, in which case our model labels the result with the highest watermark score of 1. When the watermark is small and on the edge of the image our model labels the image with a watermark score near zero. Please see the supplementary material for more results.

\section{Watermark Scoring Evaluation}
We collected 10k images for our watermark scoring dataset by displaying images to users and asking
them to score them based on the size and location of the watermarks in each image.
They were given four annotation options numbered $0$ through $3$, with $0$ being an image with no watermarks, $1$ being an image with small watermarks near its edges and $3$ being an image with large, central watermarks. We split our dataset into a training set of 8k to fit
our scoring function and a validation set of 2k to compute the MSE
of the function, which is $0.041$ for normalized score values in the range $[0,1]$. We
use the ground truth segmentation labels $L_i$ for each image as the input to our scoring function as defined in Section~\ref{sssec:score}.

\begin{figure}
\resizebox{\linewidth}{!} {
\includegraphics[height=0.5\linewidth]{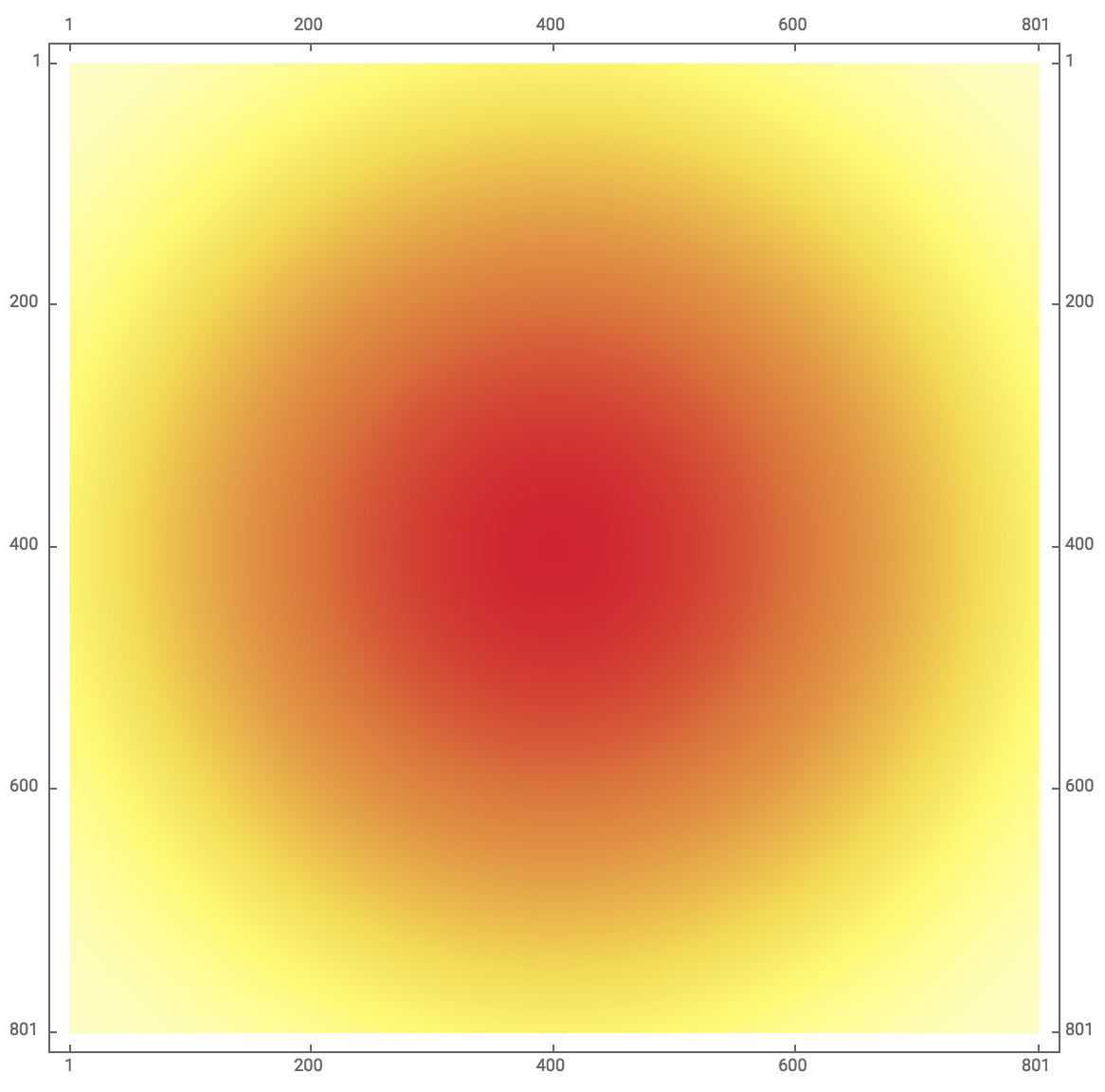}
\includegraphics[height=0.5\linewidth]{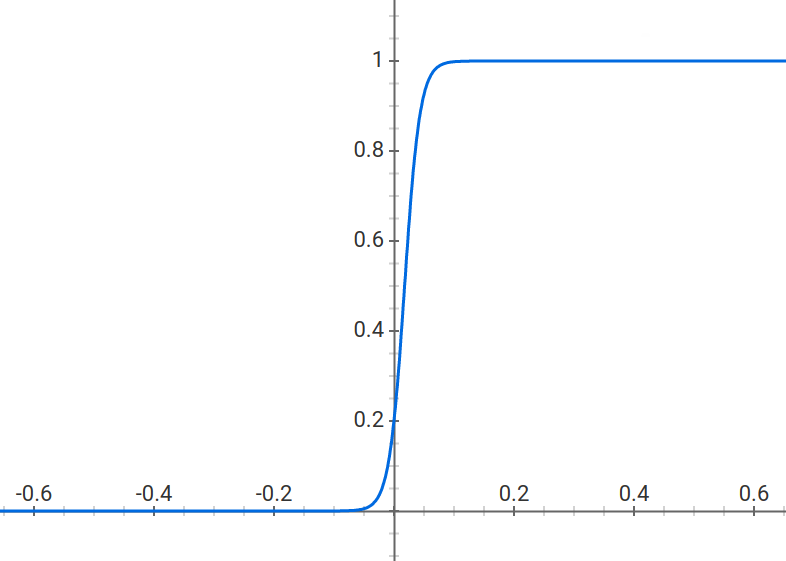}}
\caption{Visualization of the isometric 2D Gaussian (left) and the sigmoid function (right) used in our watermark scoring function by fitting them on human evaluations of watermarked images.}
\label{fig:Gaussian}
\vspace{-8pt}
\end{figure}

The optimal $\sigma$ value of our Gaussian weight function is found to be $0.44$. 
Figure~\ref{fig:Gaussian}
visualizes the Gaussian with this sigma value on the fixed image size of our dataset.
Weighting the pixel labels with this sigma value gives watermark pixels in the center
of the image a considerably larger weight than edge watermarks, with pixels near the corners
scaled by a small weight.

The optimal $\lambda$ value of our sigmoid function is found to be $78$, which results in a very steep sigmoid function, as shown in Figure~\ref{fig:Gaussian}, that is close to  a
step function. This optimal value models the human response to an increase in the area
of the watermark in an image. If a watermark takes up more than a small portion of the
image, it is given a high watermark score. For more central watermarks,
the area percentage required for a score of \textit{"very distracting"} is even lower.

With the final goal of using these watermark scores to downrank images that have obtrusive watermarks, we design an evaluation based on pairwise watermark score comparisons. We asked users to score 10k
images by the visual impact of their watermarks. These scores were obtained by displaying
an image to users and asking them to score the image based on how perceptually distracting the image's watermarks are.
They were given a range of responses from $[0,3]$, where $0$ was reserved for images that
did not contain watermarks and $3$ was for images with very distracting watermarks.
Each image was displayed to three users, and their responses are averaged to obtain the image's final score.
This score represents how large of an effect the image's watermarks had on the image's perceptual quality. To evaluate our model's image ranking against
the ground truth, we compute the accuracy of our ranking for all potential image
pairs. In Table~\ref{table:ScoreGroundTruthPrecision}, we show the percentage
of image pairs $(A,B)$ with true ranking $A > B$, whose predicted score $\hat{A} > \hat{B}$. For instance, when all the images with score of $3$ are compared against all the images with score of $1$, in 95.36\% of those pairs images with ground truth score of $3$ have higher watermark score predicted by our model.
In other words, our evaluation determines that for a quality ranking task our model will rank images with very distracting watermarks lower than images with ignorable watermarks ~95\% of the time.

In Table~\ref{table:ScorePredictedLabelsPositive}, we use the predicted segmentation
labels from our hybrid model as input to our sigmoid function. In this experiment,
we eliminate images that we predict to not contain watermarks, as these images
are not given a watermark score. Our model's ranking of images with very
distracting watermarks against images with ignorable watermarks had an accuracy
of 96\%. In addition, our ranking accuracy of image pairs for all validation images is shown in Table~\ref{table:ScorePredictedLabels}.

\begin{table}
\centering
\begin{tabular}{|c|c|c|c|c|c|}
\hline
 & 3 & 2 & 1 \\
\hline\hline
0 & 96.71 & 96.45 & 92.08 \\
\hline
1 & 95.36 & 81.05 & N/A \\
\hline
2 & 88.25 & N/A & N/A \\
\hline
\end{tabular}
\caption{Percentage pairwise accuracy for image pairs with different ground truth scores shown in the first row and the column. Evaluated on 10k images.}
\label{table:ScoreGroundTruthPrecision}
\vspace{-8pt}
\end{table}

\begin{table}
\centering
\begin{tabular}{|c|c|c|c|c|c|}
\hline
 & 3 & 2 & 1 \\
\hline\hline
0 & N/A & N/A & N/A\\
\hline
1 & 96.14 & 78.39 & N/A\\
\hline
2 & 87.80 & N/A & N/A\\
\hline
\end{tabular}
\caption{Percentage pairwise accuracy for image pairs with different ground truth scores shown in the first row and the column. Evaluated on 1k randomly sampled images from the validation set, with negative predictions removed.}
\label{table:ScorePredictedLabelsPositive}
\vspace{-8pt}
\end{table}

\begin{table}
\centering
\begin{tabular}{|c|c|c|c|c|c|}
\hline
 & 3 & 2 & 1 \\
\hline\hline
0 & 86.76 & 81.10 & 57.95 \\
\hline
1 & 86.40 & 72.42 & N/A \\
\hline
2 & 79.47 & N/A & N/A\\
\hline
\end{tabular}
\caption{Percentage pairwise accuracy for image pairs with different ground truth scores shown in the first row and the column. Evaluated on 1.7k randomly sampled images from the validation set.}
\label{table:ScorePredictedLabels}
\vspace{-8pt}
\end{table}

\section{Discussion and Future Work}

We proposed a model to detect watermarks on images that generalizes to real world datasets, without
any synthetic data. We consider our image scoring
problem in the context of sparse positive examples with high variance. Previous
work has produced interesting results for watermark detection, but these cases have
been limited primarily to synthetic data with an evaluation set of select watermarks that
were used to train the model. In our work, we focus on inferring a score for
watermarked images that is accurate in the context of large-scale image ranking, which requires a generalized
model that can recognize watermarks not present in our training set.

A potential future direction for watermark scoring is implementing a multi-class model.
For this model, watermark pixels in images would be labelled based on their
opacity in the image. This information could then be used in our scoring function
to scale the weight of the predicted watermark pixels. Through our evaluation
of our scoring function, a high accuracy can still be attained by only considering
the segmentation of images into watermark and non-watermark regions.

\fboxsep=0pt
\fboxrule=1pt
\begin{table*}
\resizebox{2\columnwidth}{!} {
\begin{tabular}{ccc|ccc}
\hline
Original Image & Segmentation & Score &Original Image & Segmentation & Score \\
\hline 
\includegraphics[width=0.25\textwidth]{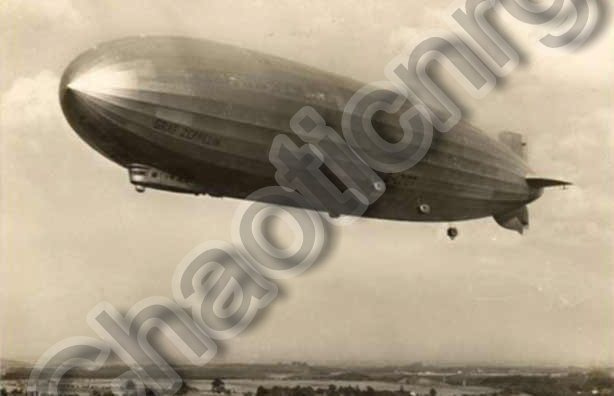} 
&
\fbox{\includegraphics[width=0.25\textwidth]{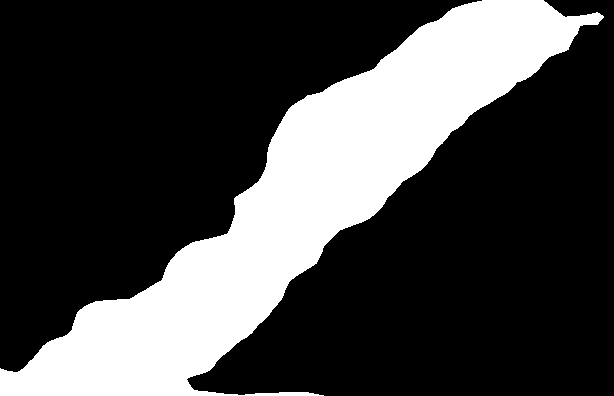}} 
& 
1.0
&
\includegraphics[width=0.25\textwidth]{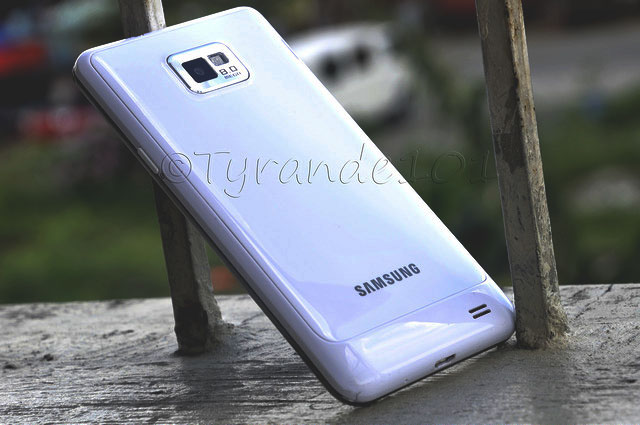}
&
\fbox{\includegraphics[width=0.25\textwidth]{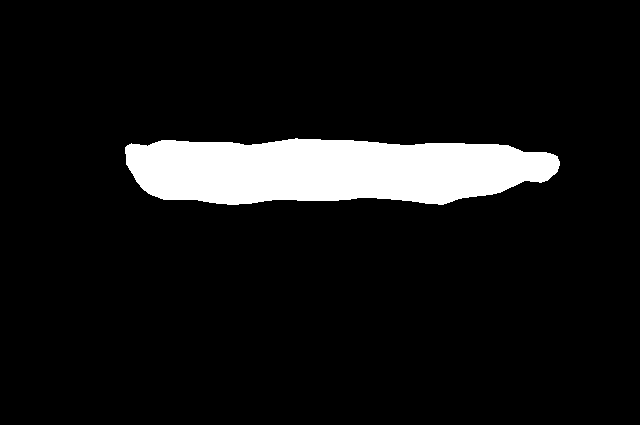}}
& 
0.998
\\
\includegraphics[width=0.25\textwidth]{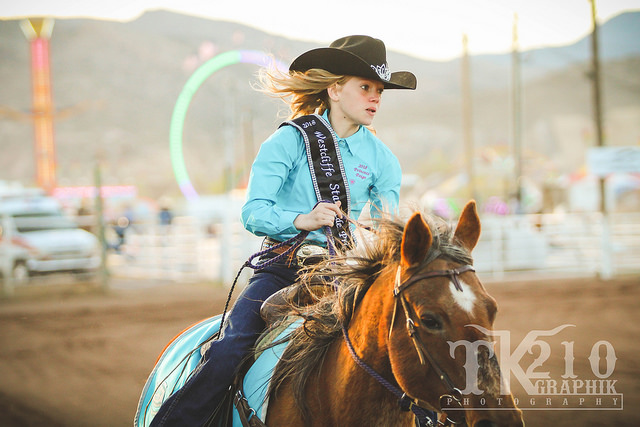}
&
\fbox{\includegraphics[width=0.25\textwidth]{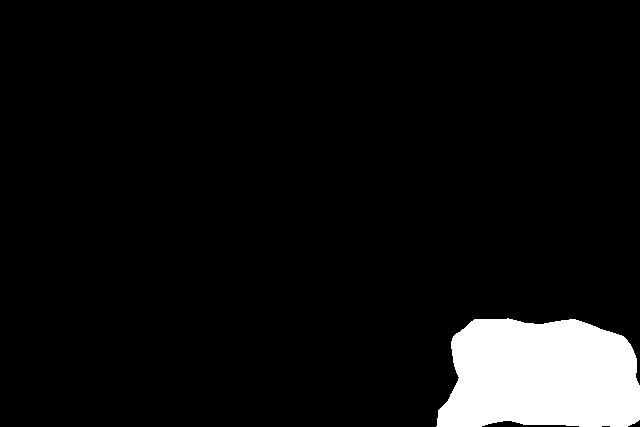}}
& 
0.961
&
\includegraphics[width=0.25\textwidth]{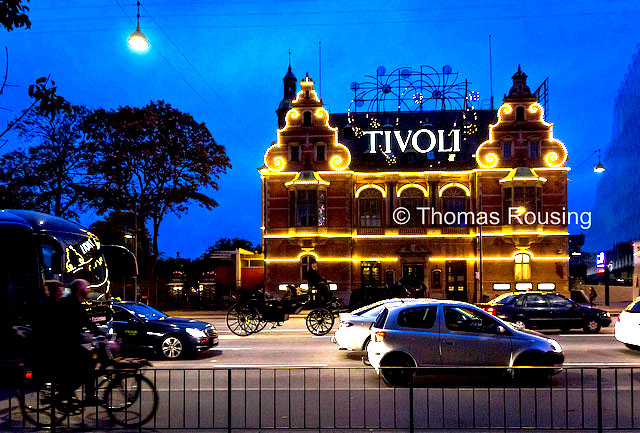}
&
\fbox{\includegraphics[width=0.25\textwidth]{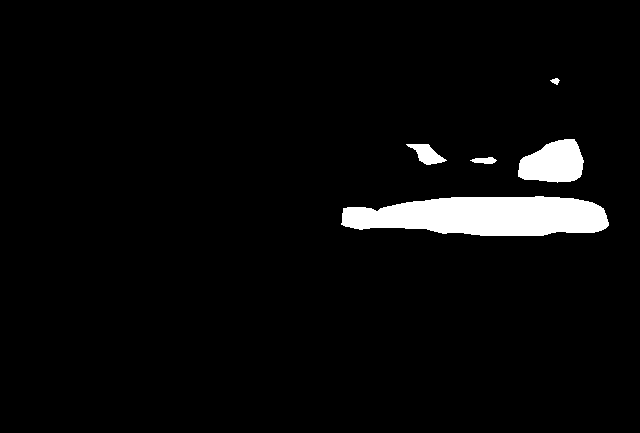}}
& 
0.840
\\
\includegraphics[width=0.25\textwidth]{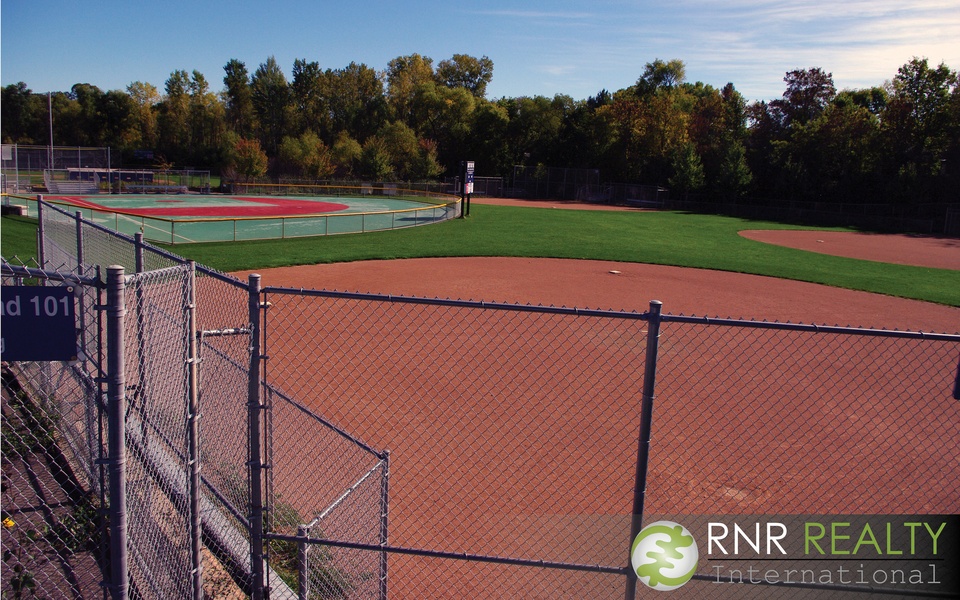}
&
\fbox{\includegraphics[width=0.25\textwidth]{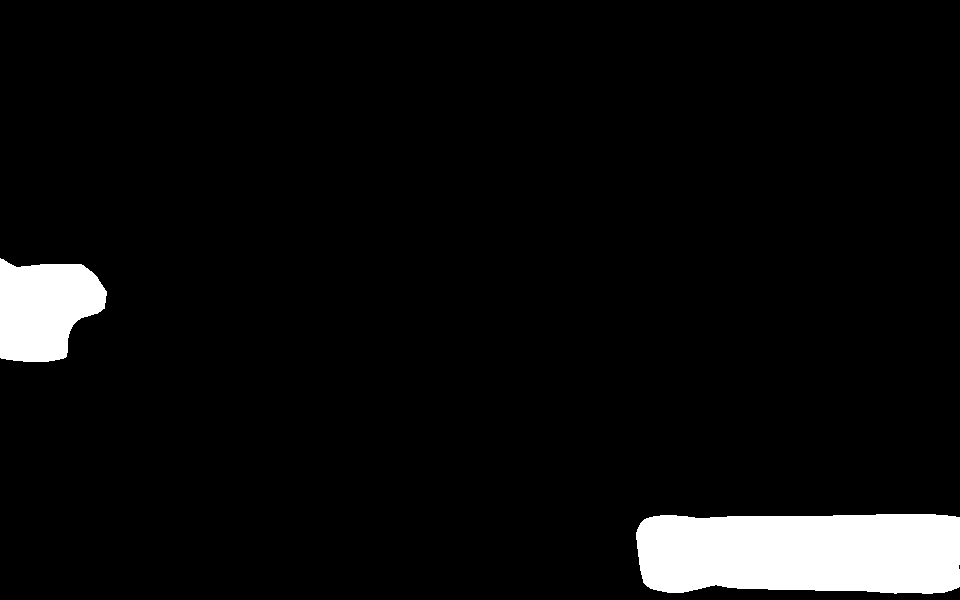}}
& 
0.831
&
\includegraphics[width=0.25\textwidth]{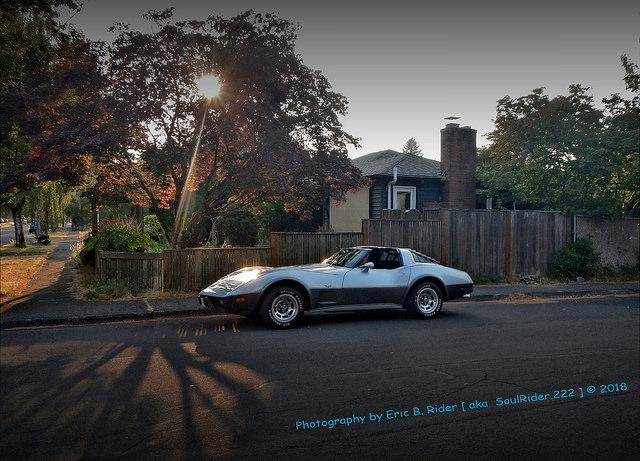}
&
\fbox{\includegraphics[width=0.25\textwidth]{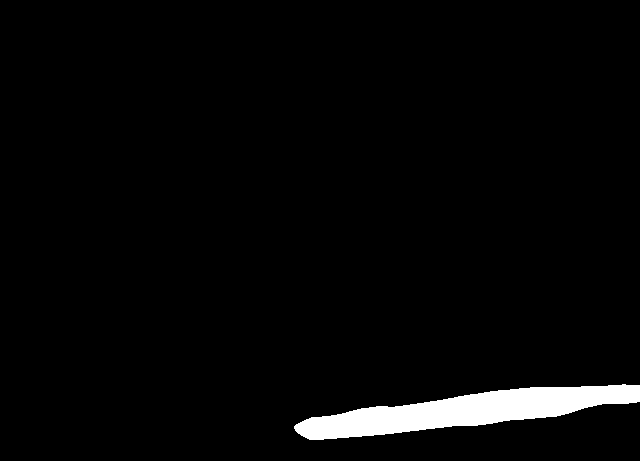}}
& 
0.641
\\
\includegraphics[width=0.25\textwidth]{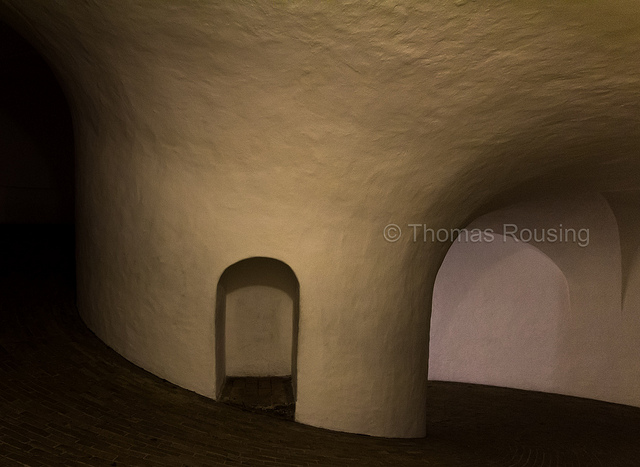} 
&
\fbox{\includegraphics[width=0.25\textwidth]{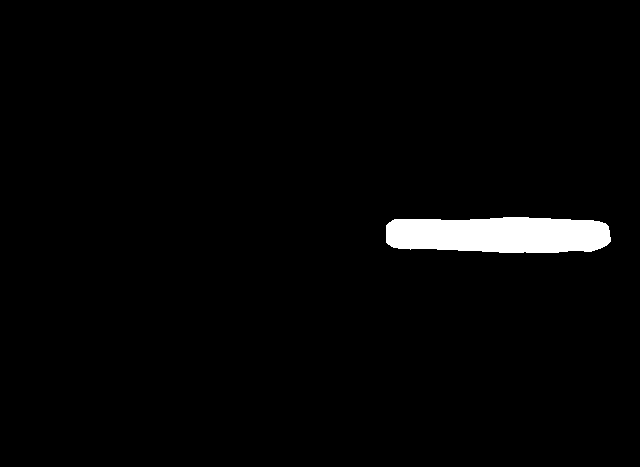}}
& 
0.553
&
\includegraphics[width=0.25\textwidth]{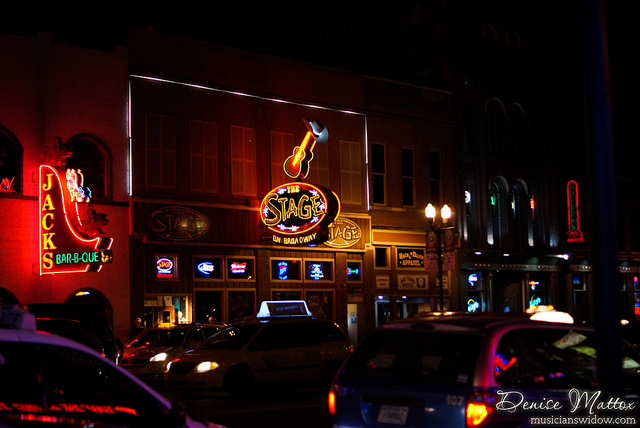}
&
\fbox{\includegraphics[width=0.25\textwidth]{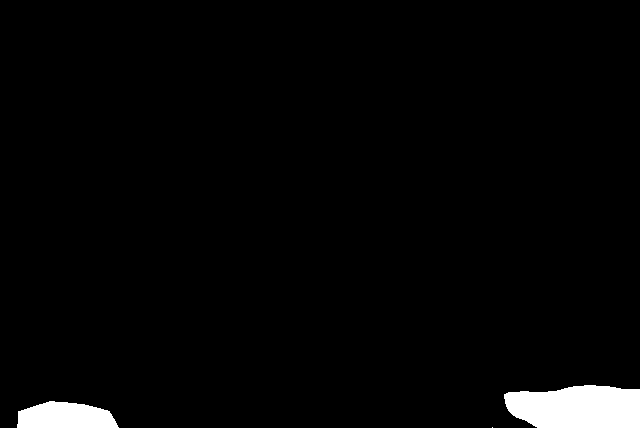}}
& 
0.451
\\
\includegraphics[width=0.25\textwidth]{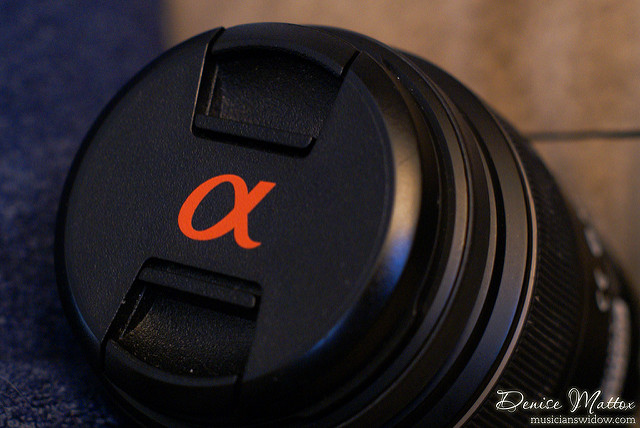}
&
\fbox{\includegraphics[width=0.25\textwidth]{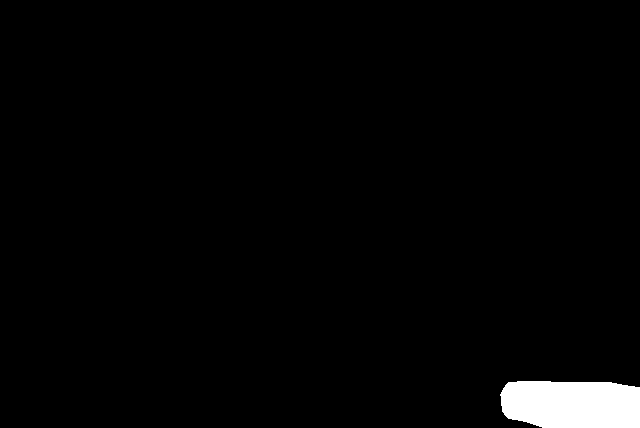}}
& 
0.379
&
\includegraphics[width=0.25\textwidth]{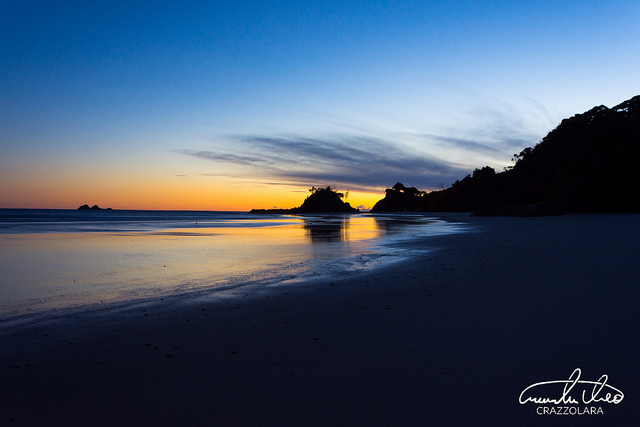}
&
\fbox{\includegraphics[width=0.25\textwidth]{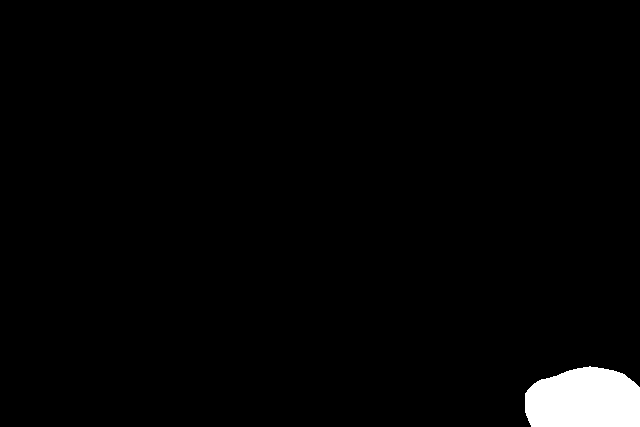}}
& 
0.370
\\
\includegraphics[width=0.25\textwidth]{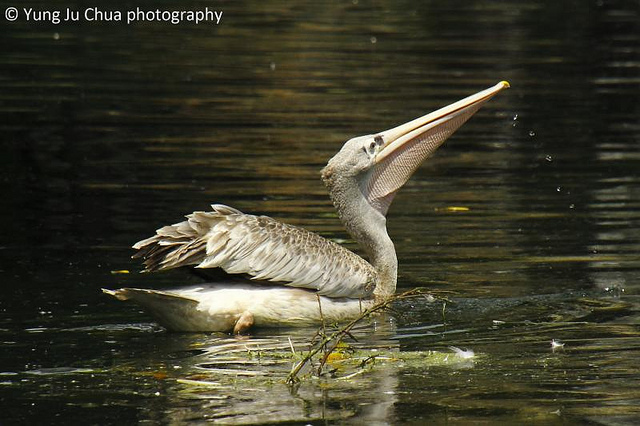}
&
\fbox{\includegraphics[width=0.25\textwidth]{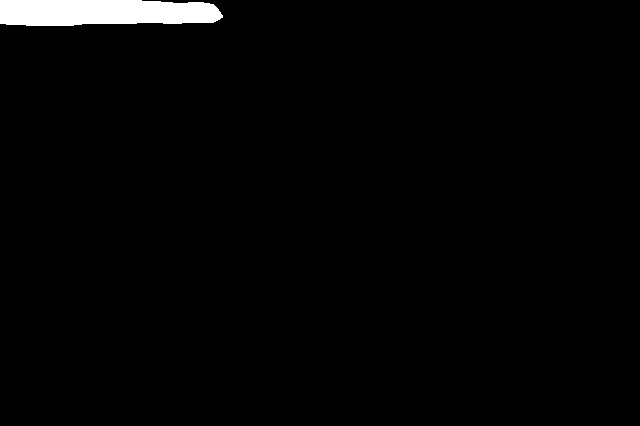}}
& 
0.333
&
\includegraphics[width=0.25\textwidth]{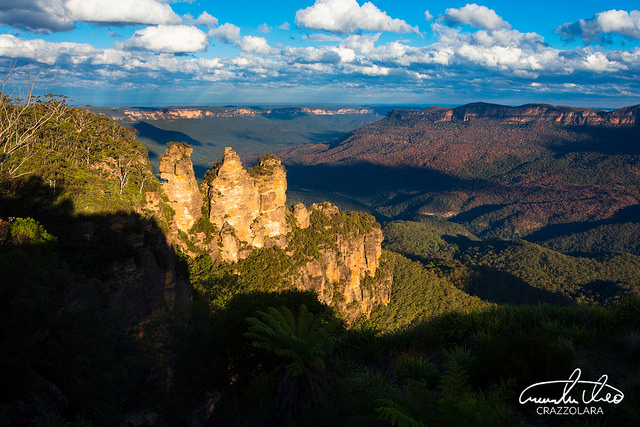}
&
\fbox{\includegraphics[width=0.25\textwidth]{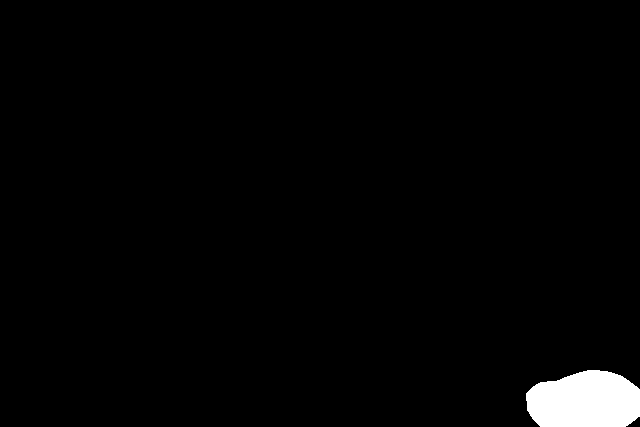}}
& 
0.328
\\
\includegraphics[width=0.25\textwidth]{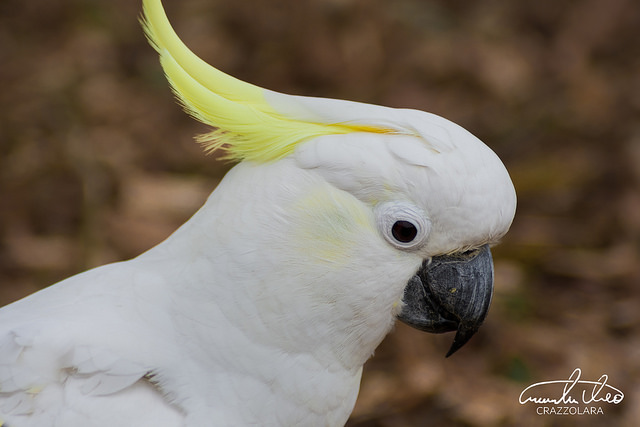}
&
\fbox{\includegraphics[width=0.25\textwidth]{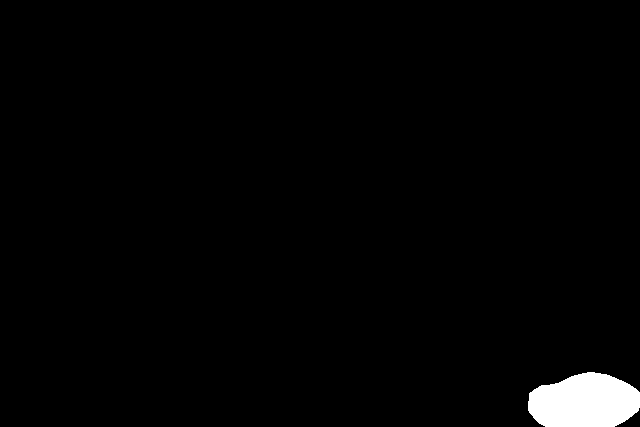}}
& 
0.292
&
\includegraphics[width=0.25\textwidth]{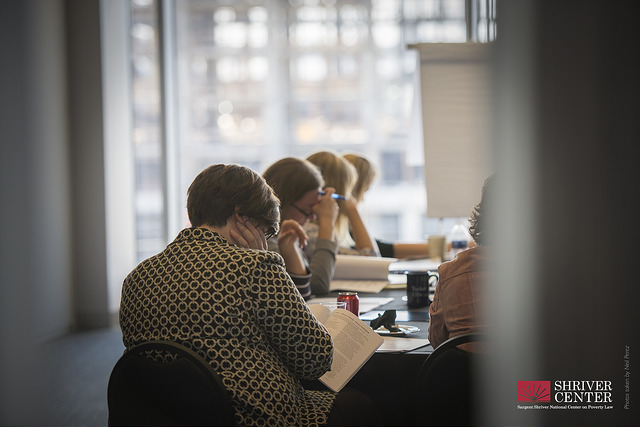}
&
\fbox{\includegraphics[width=0.25\textwidth]{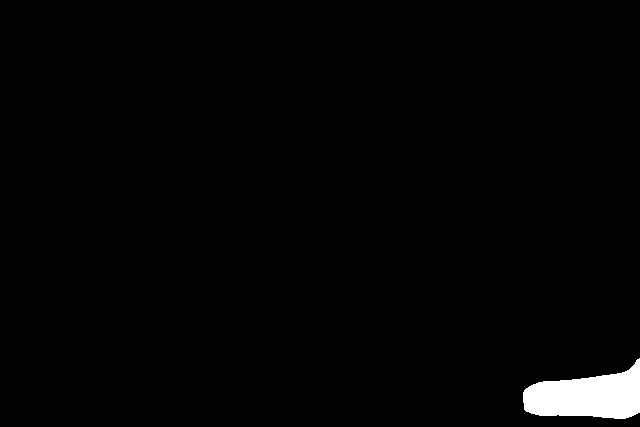}}
& 
0.266
\\
\includegraphics[width=0.25\textwidth]{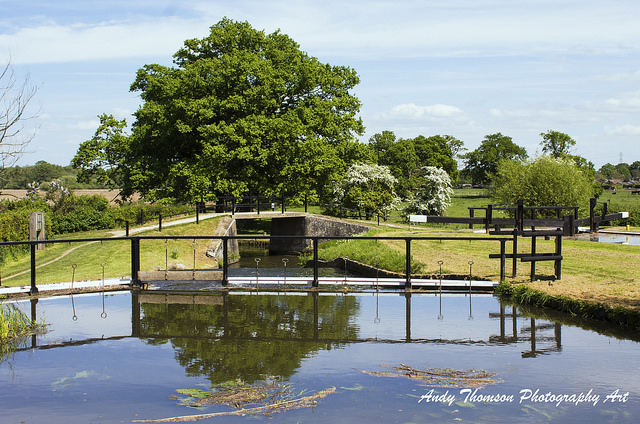}
&
\fbox{\includegraphics[width=0.25\textwidth]{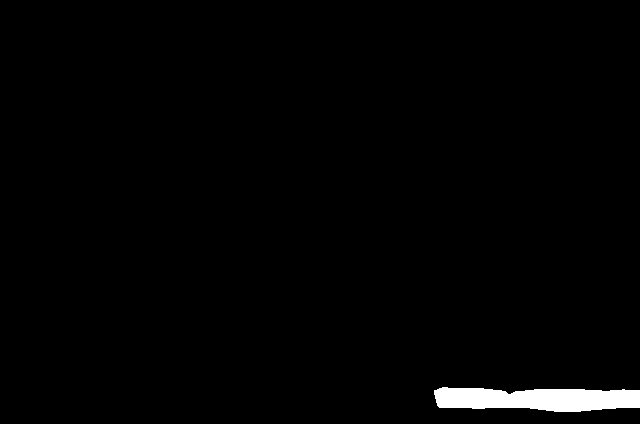}}
& 
0.257
&
\includegraphics[width=0.25\textwidth]{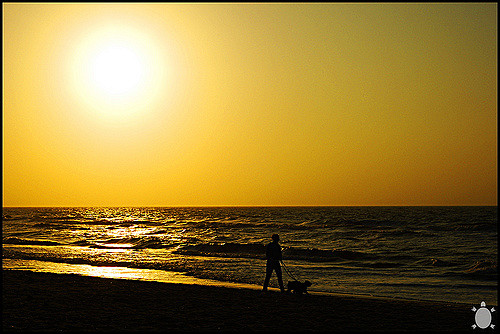}
&
\fbox{\includegraphics[width=0.25\textwidth]{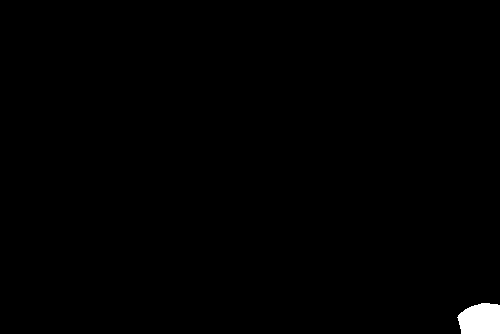}}
& 
0.102
\\
\includegraphics[width=0.25\textwidth]{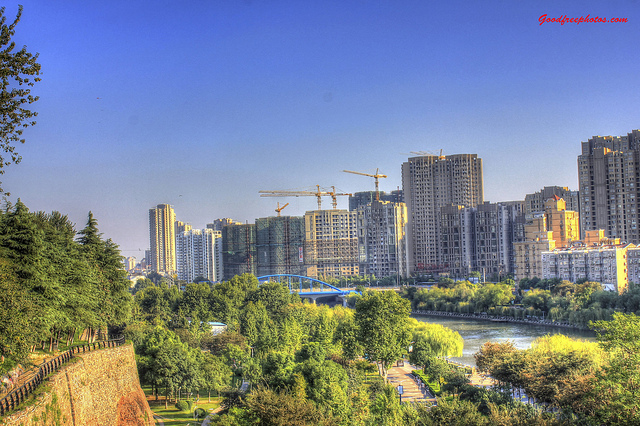}
&
\fbox{\includegraphics[width=0.25\textwidth]{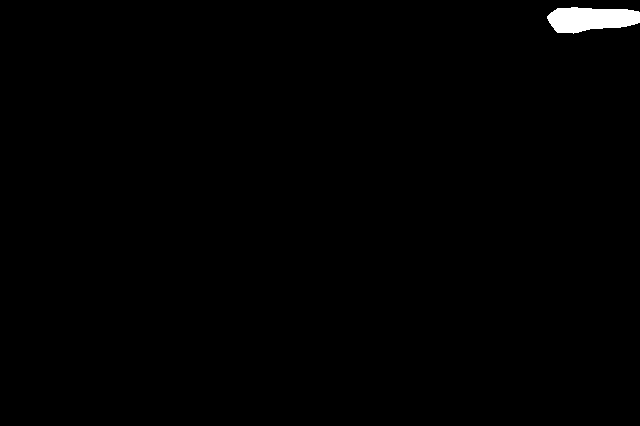}}
& 
0.093
&
\includegraphics[width=0.25\textwidth]{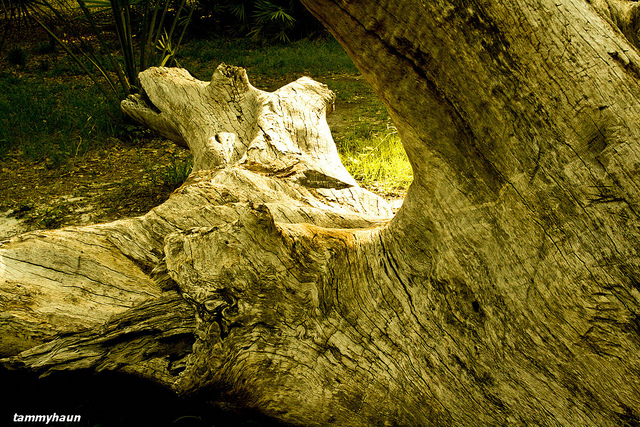}
&
\fbox{\includegraphics[width=0.25\textwidth]{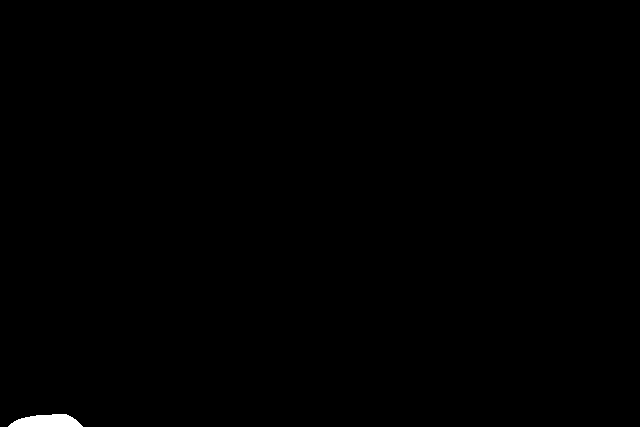}}
& 
0.033
\\
\hline
\end{tabular}}
\caption{Watermark detection results on Flickr images~\cite{Flickr} that were not included in either training or testing. From left to right: original image, detected watermark segmentation, and final watermark score between [0, 1]. Please note that some watermarks are subtle to see in a printout, refer to the supplemental material for higher resolution.}
\label{table:FlickrTable}
\end{table*}

{\small
\bibliographystyle{ieee}
\bibliography{egbib}
}

\pagebreak

\onecolumn

\fboxsep=0pt
\fboxrule=1pt

\begin{table*}[h]
\caption{More watermark detection results on Flickr images~\cite{Flickr} that were not included in training or testing. From left to right: original image, detected watermark segmentation, and final watermark score between [0, 1]. Images are ordered by decreasing watermark scores.}
\vspace{20pt}
\begin{tabular}{ccc}
\hline
Original Image & Segmentation & Watermark Score \\
\hline
\includegraphics[width=0.35\textwidth]{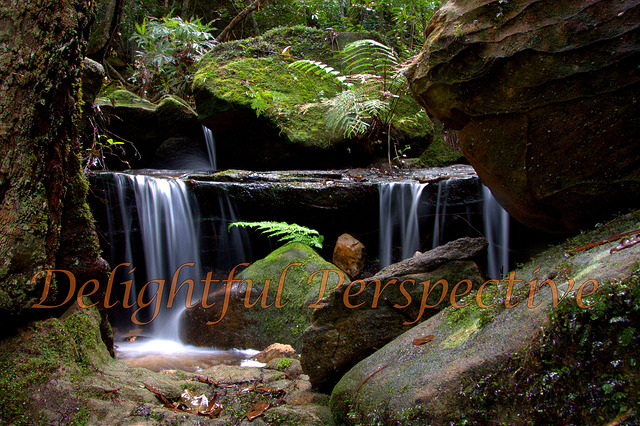} &
\fbox{\includegraphics[width=0.35\textwidth]{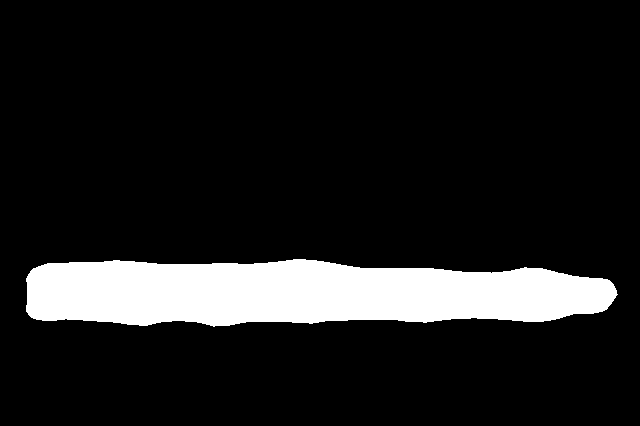}} & 1.0 \\
\includegraphics[width=0.35\textwidth]{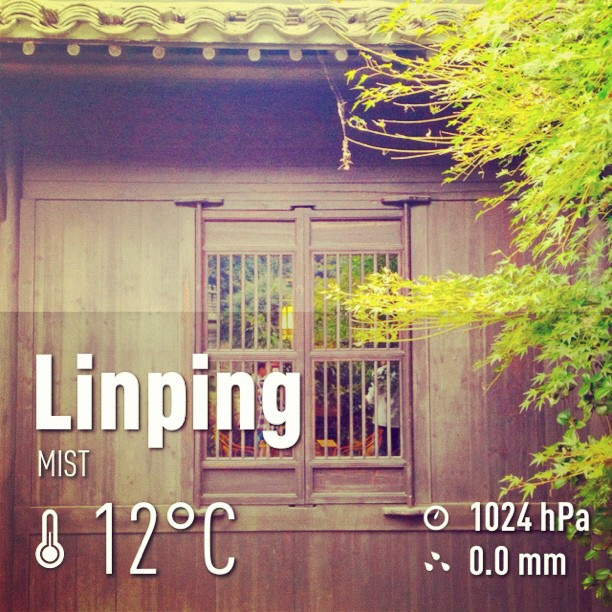} &
\fbox{\includegraphics[width=0.35\textwidth]{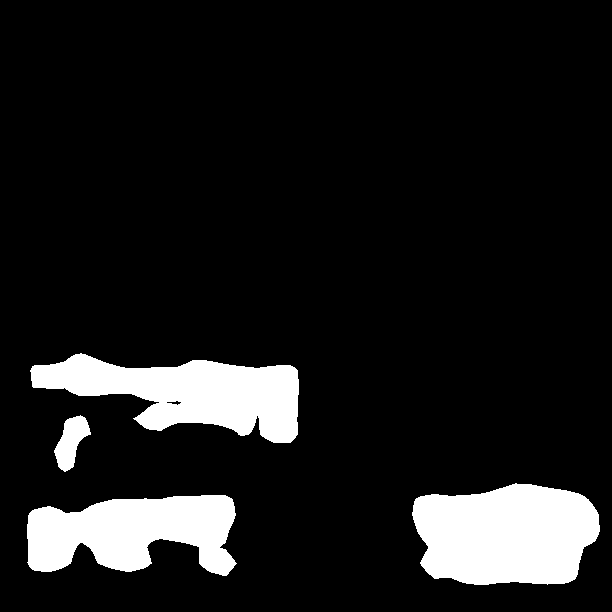}} & 1.0 \\
\includegraphics[width=0.35\textwidth]{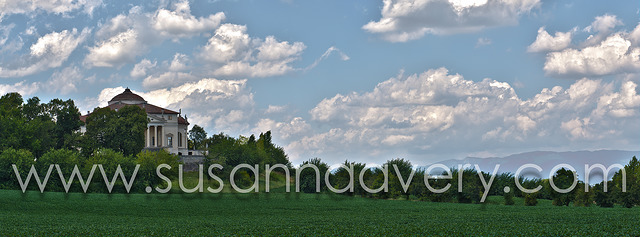} &
\fbox{\includegraphics[width=0.35\textwidth]{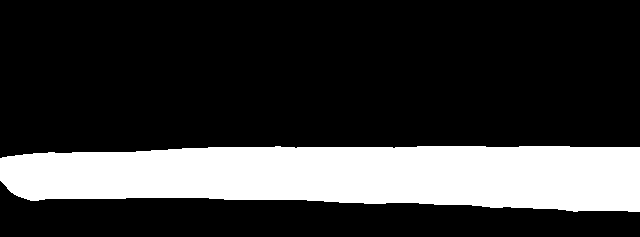}} & 1.0 \\
\hline
\end{tabular}
\end{table*}

\begin{table*}
\begin{tabular}{ccc}
\hline
Original Image & Segmentation & Watermark Score \\
\hline
\includegraphics[width=0.35\textwidth]{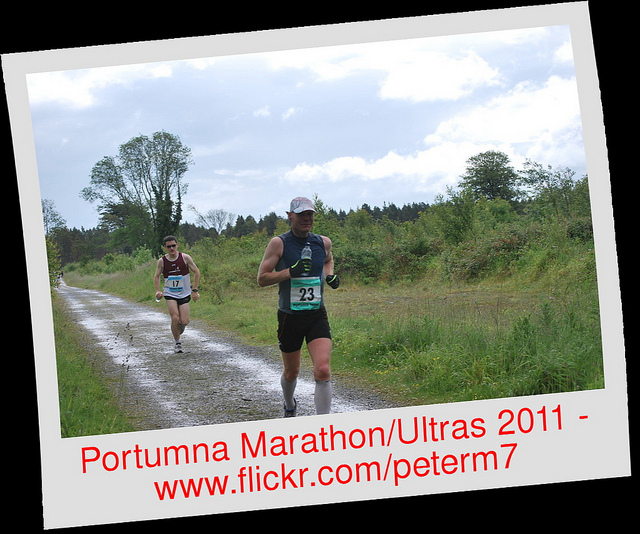} &
\fbox{\includegraphics[width=0.35\textwidth]{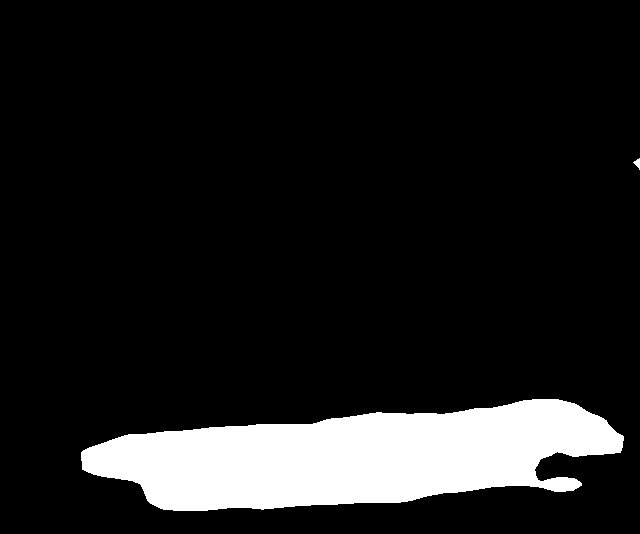}} & 1.0 \\
\includegraphics[width=0.35\textwidth]{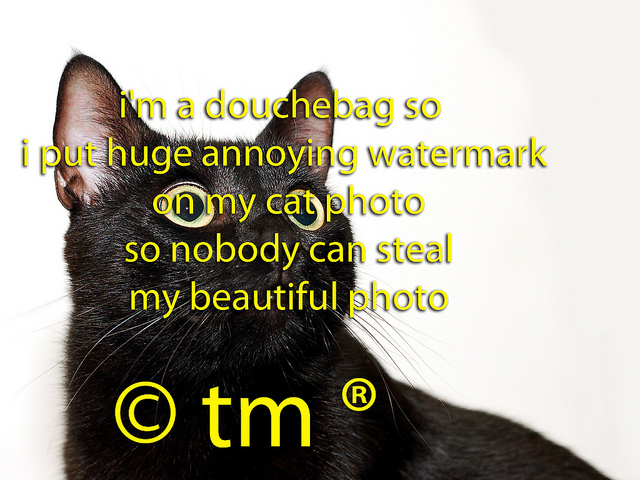} &
\fbox{\includegraphics[width=0.35\textwidth]{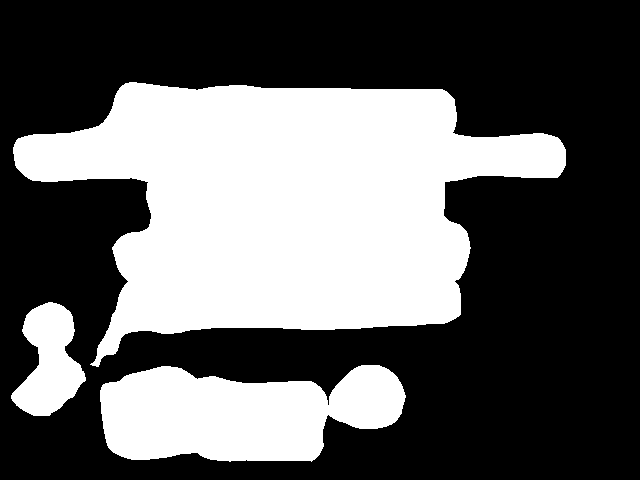}} & 1.0 \\
\includegraphics[width=0.35\textwidth]{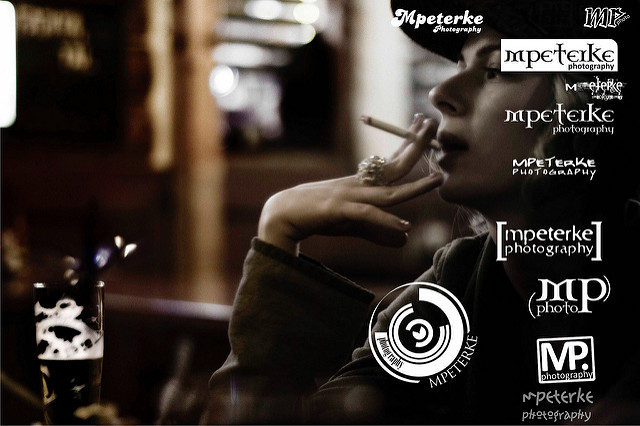} &
\fbox{\includegraphics[width=0.35\textwidth]{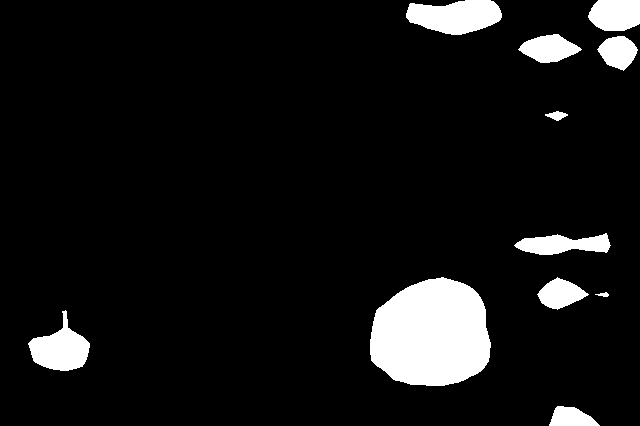}} & 0.989 \\
\includegraphics[width=0.35\textwidth]{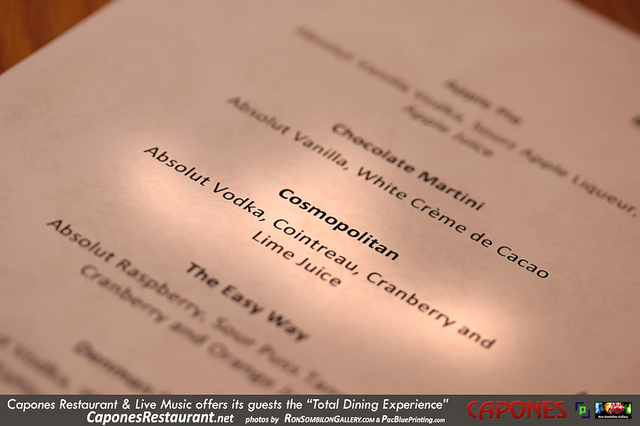} &
\fbox{\includegraphics[width=0.35\textwidth]{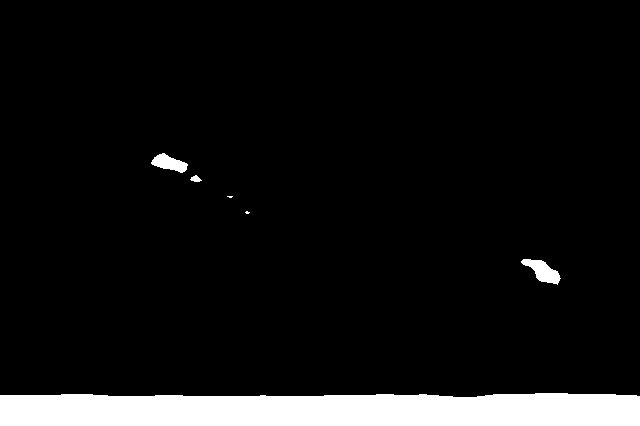}} & 0.980 \\
\includegraphics[width=0.35\textwidth]{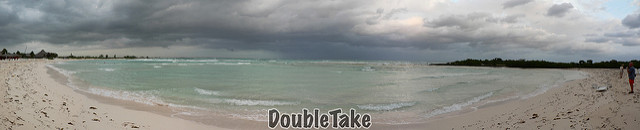} &
\fbox{\includegraphics[width=0.35\textwidth]{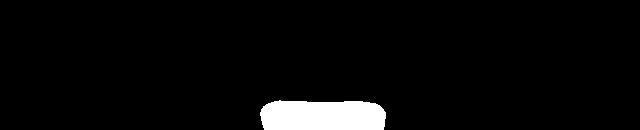}} & 0.883 \\
\includegraphics[width=0.35\textwidth]{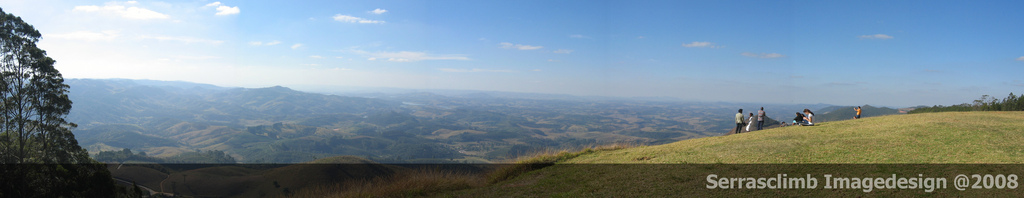} &
\fbox{\includegraphics[width=0.35\textwidth]{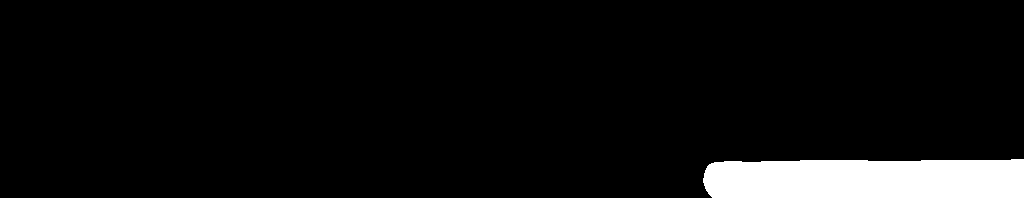}} & 0.877 \\
\hline
\end{tabular}
\end{table*}

\begin{table*}
\begin{tabular}{ccc}
\hline
Original Image & Segmentation & Watermark Score \\
\hline
\includegraphics[width=0.35\textwidth]{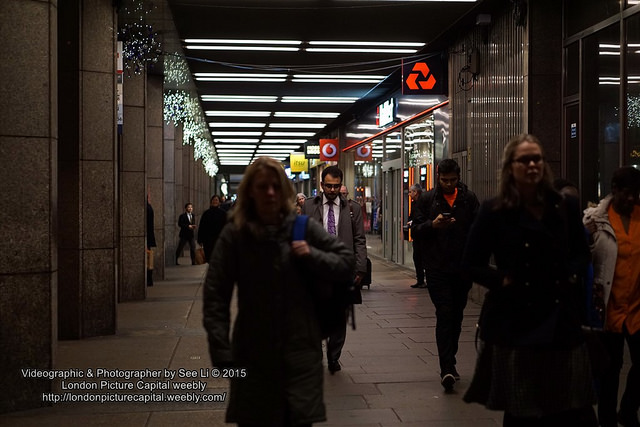} &
\fbox{\includegraphics[width=0.35\textwidth]{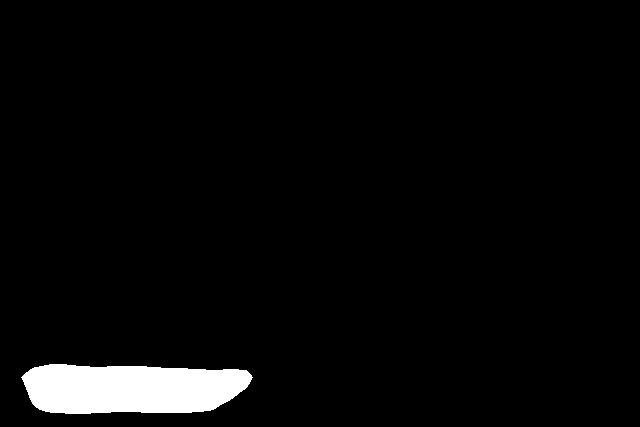}} & 0.694 \\
\includegraphics[width=0.35\textwidth]{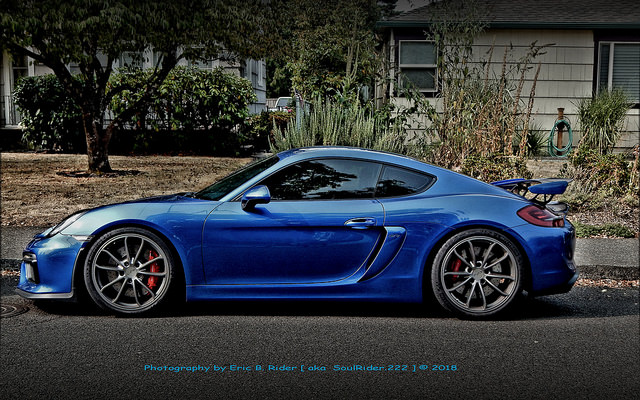} &
\fbox{\includegraphics[width=0.35\textwidth]{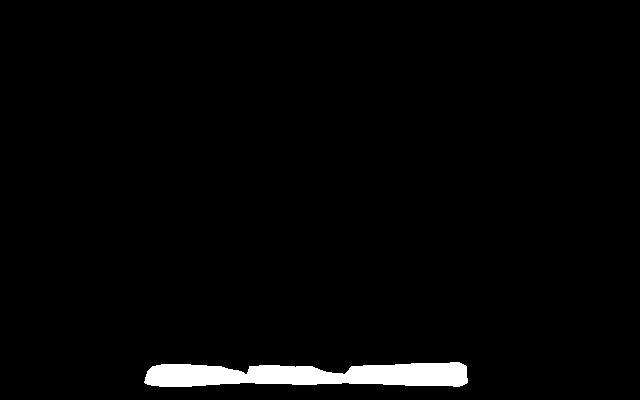}} & 0.536 \\
\includegraphics[width=0.35\textwidth]{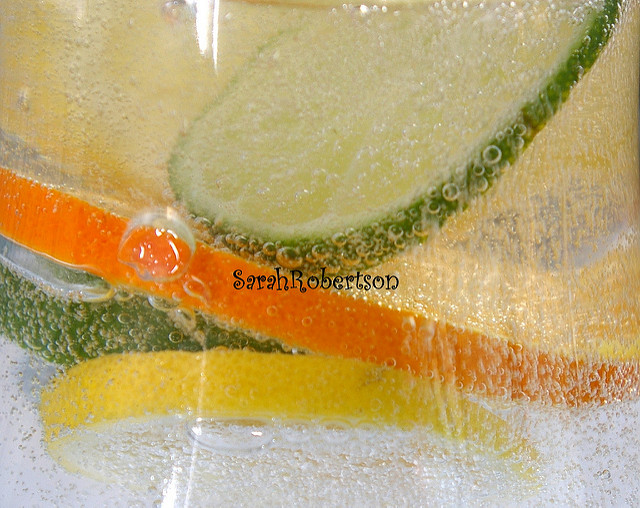} &
\fbox{\includegraphics[width=0.35\textwidth]{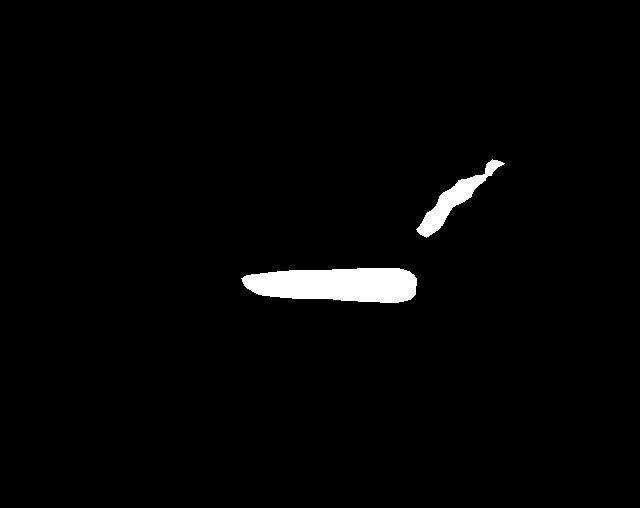}} & 0.533 \\
\includegraphics[width=0.35\textwidth]{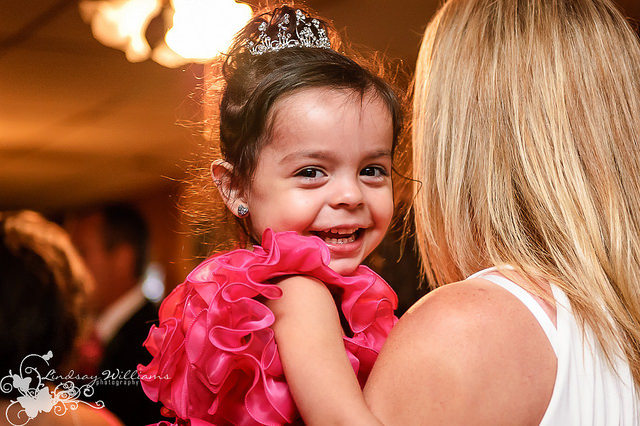} &
\fbox{\includegraphics[width=0.35\textwidth]{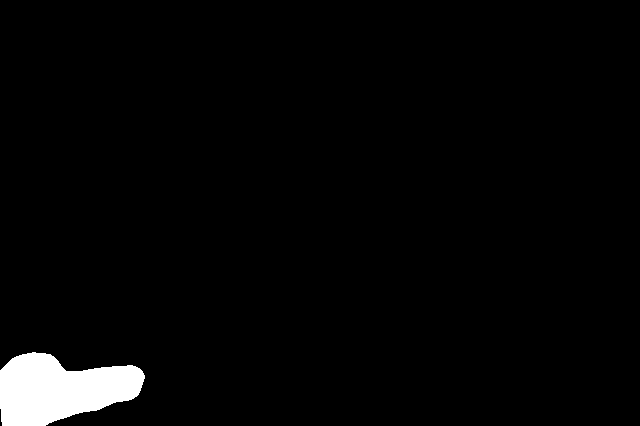}} & 0.478 \\
\includegraphics[width=0.35\textwidth]{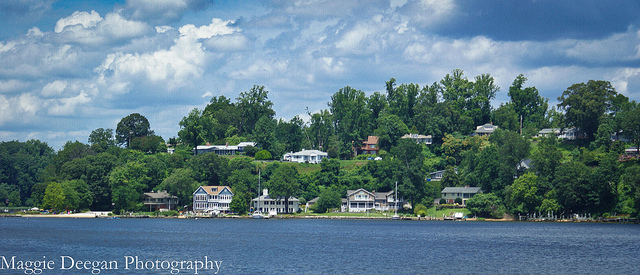} &
\fbox{\includegraphics[width=0.35\textwidth]{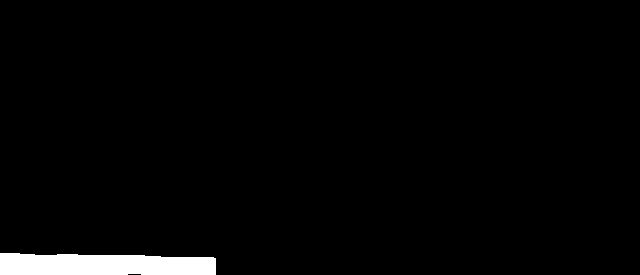}} & 0.474 \\
\hline
\end{tabular}
\end{table*}

\begin{table*}
\begin{tabular}{ccc}
\hline
Original Image & Segmentation & Watermark Score \\
\hline
\includegraphics[width=0.35\textwidth]{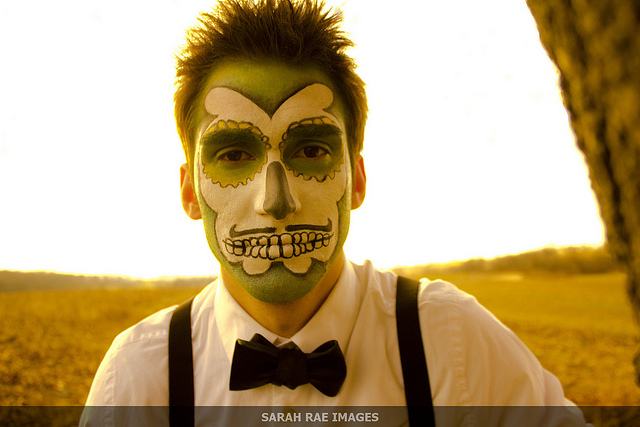} &
\fbox{\includegraphics[width=0.35\textwidth]{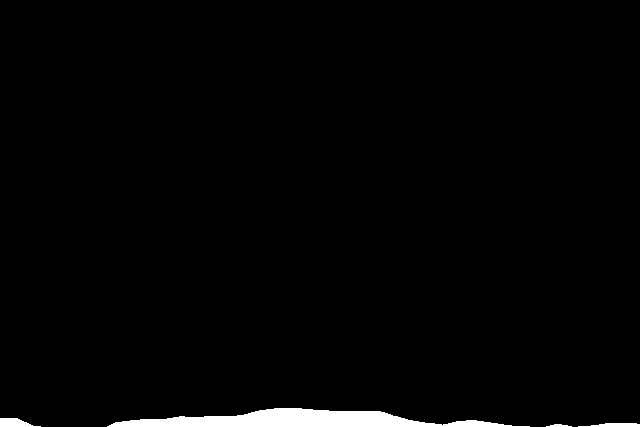}} & 0.356 \\
\includegraphics[width=0.35\textwidth]{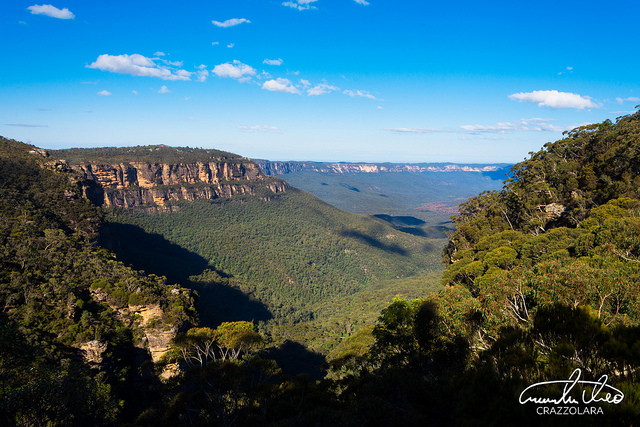} &
\fbox{\includegraphics[width=0.35\textwidth]{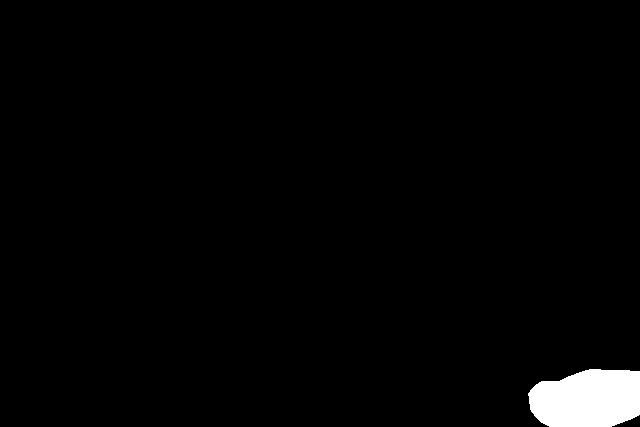}} & 0.330 \\
\includegraphics[width=0.35\textwidth]{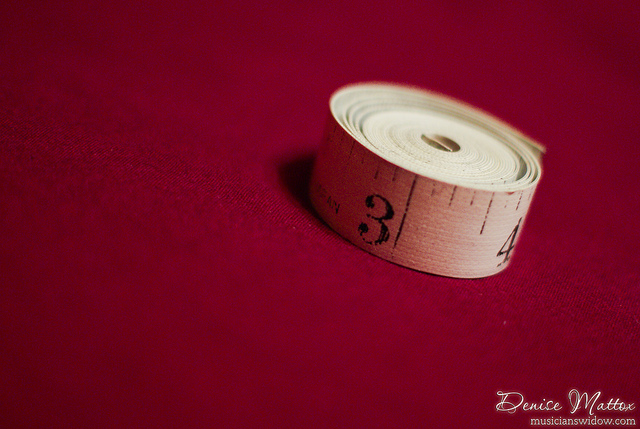} &
\fbox{\includegraphics[width=0.35\textwidth]{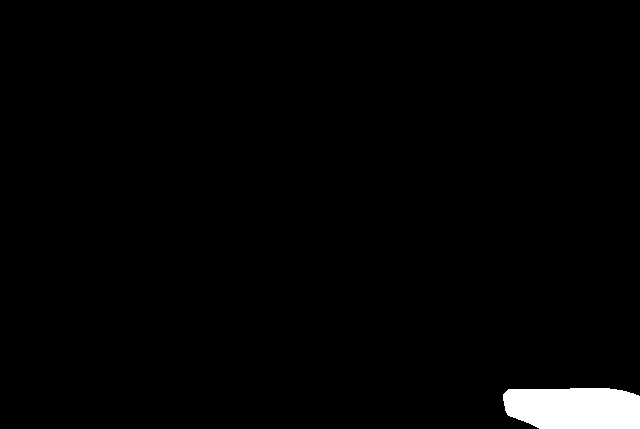}} & 0.308 \\
\includegraphics[width=0.35\textwidth]{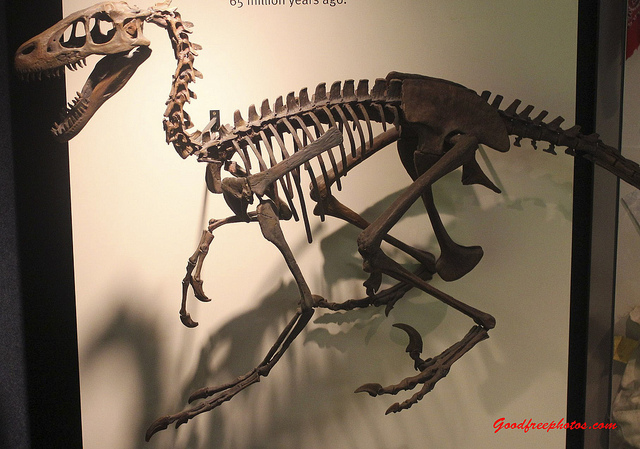} &
\fbox{\includegraphics[width=0.35\textwidth]{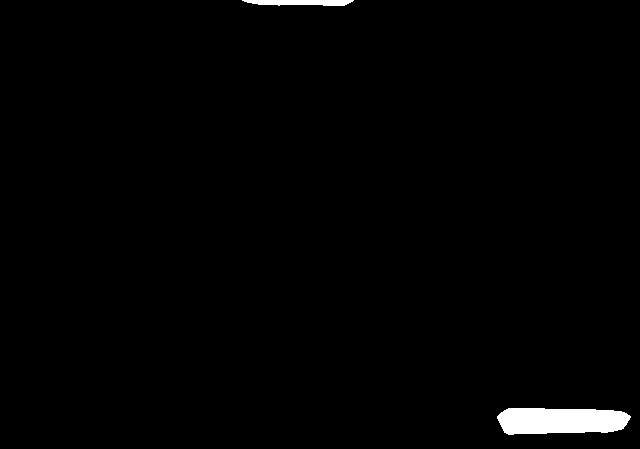}} & 0.204 \\
\includegraphics[width=0.35\textwidth]{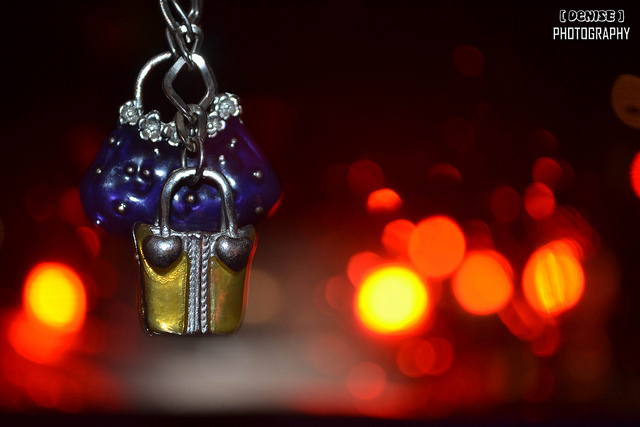} &
\fbox{\includegraphics[width=0.35\textwidth]{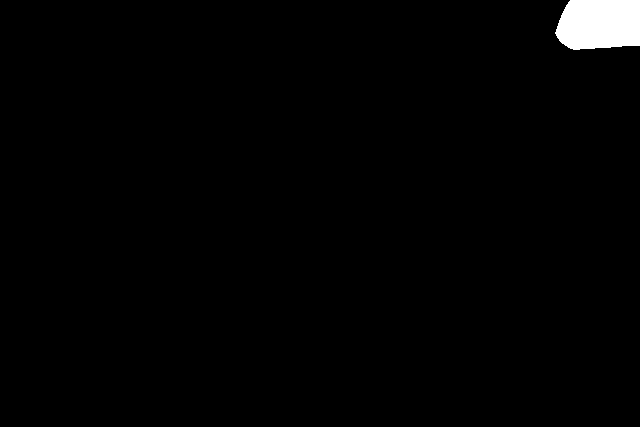}} & 0.206 \\
\hline
\end{tabular}
\end{table*}

\begin{table*}
\begin{tabular}{ccc}
\hline
Original Image & Segmentation & Watermark Score \\
\hline
\includegraphics[width=0.35\textwidth]{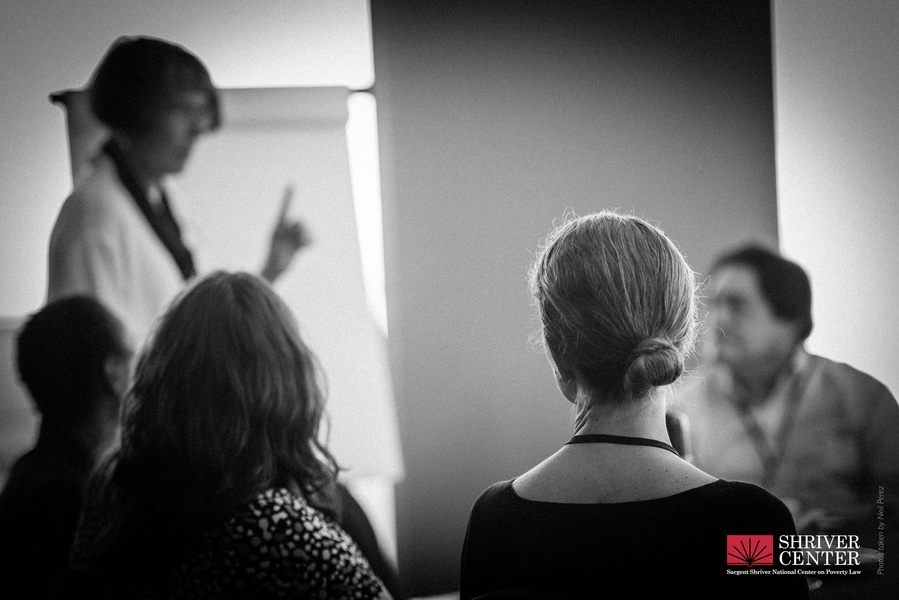} &
\fbox{\includegraphics[width=0.35\textwidth]{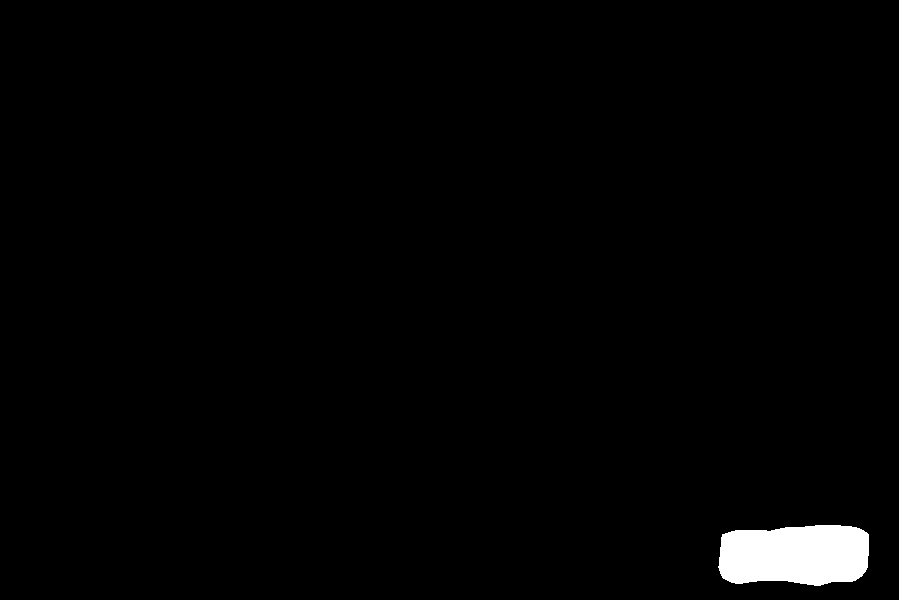}} & 0.195 \\
\includegraphics[width=0.35\textwidth]{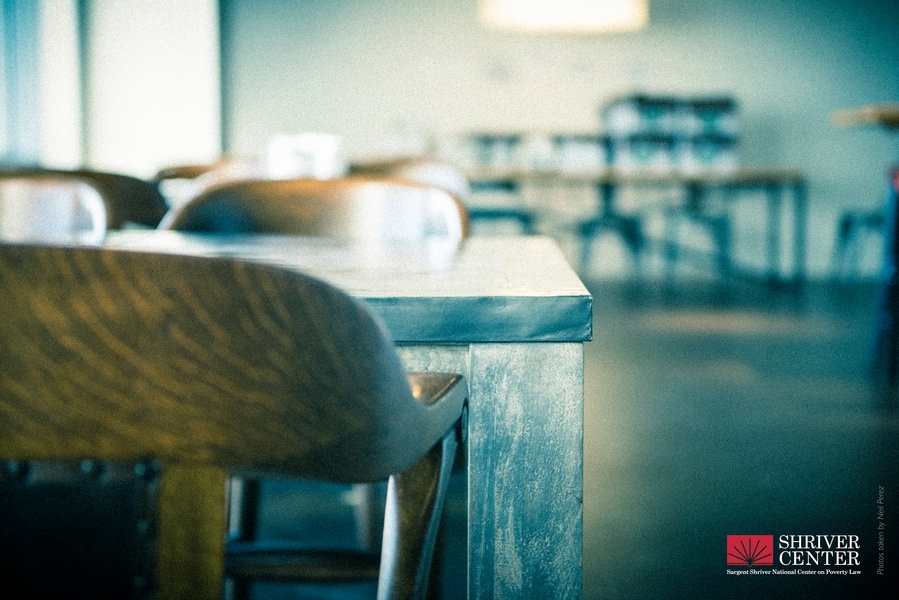} &
\fbox{\includegraphics[width=0.35\textwidth]{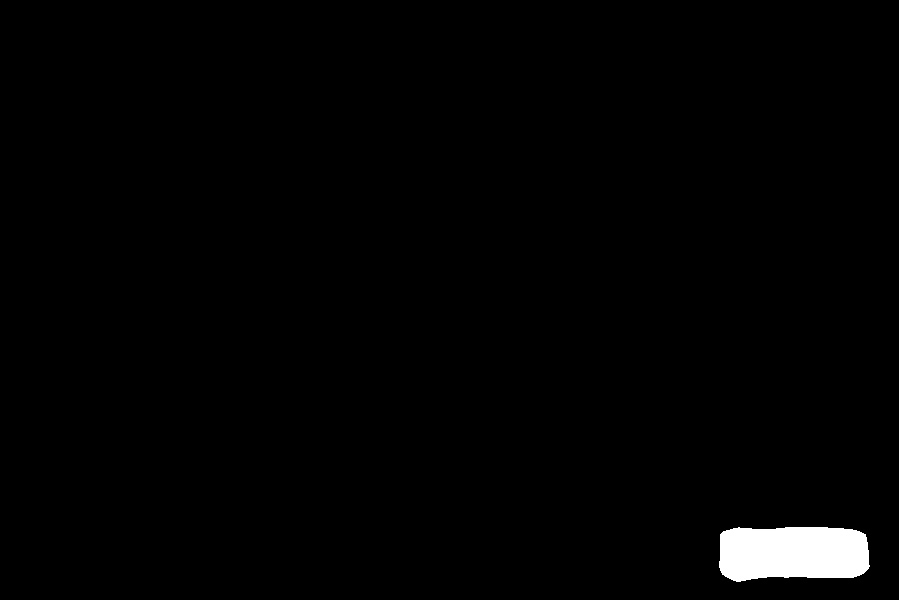}} & 0.178 \\
\includegraphics[width=0.35\textwidth]{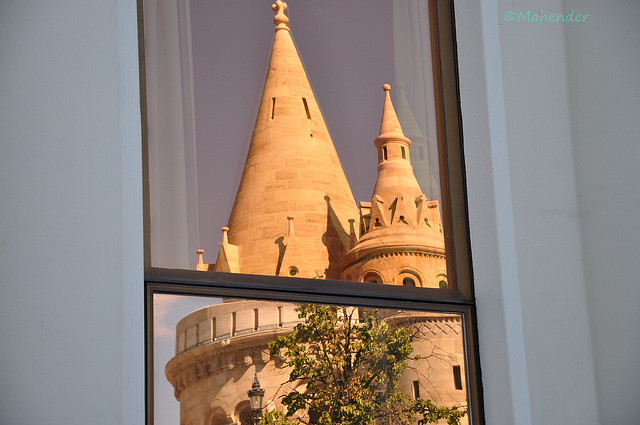} &
\fbox{\includegraphics[width=0.35\textwidth]{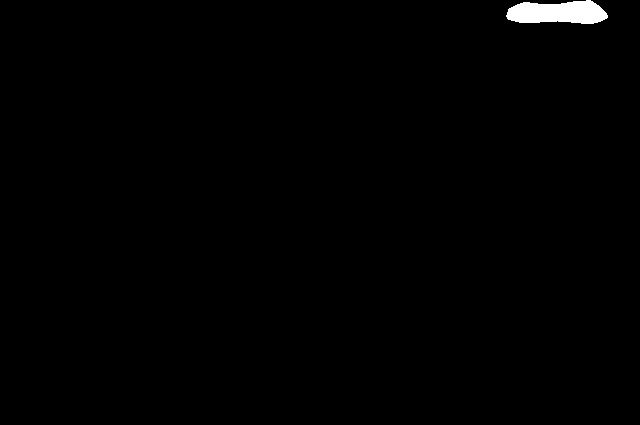}} & 0.101 \\
\includegraphics[width=0.35\textwidth]{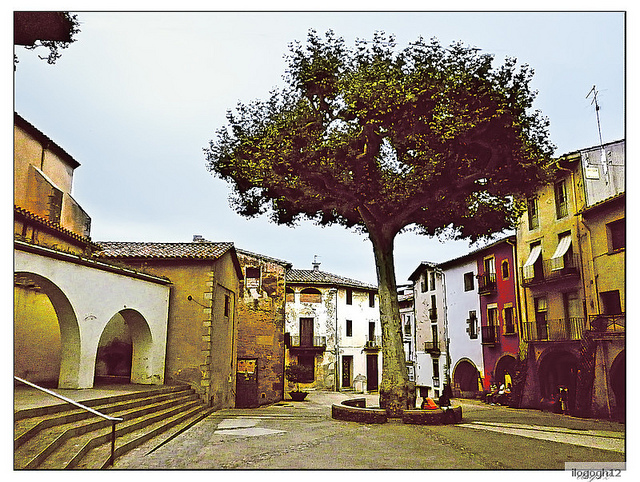} &
\fbox{\includegraphics[width=0.35\textwidth]{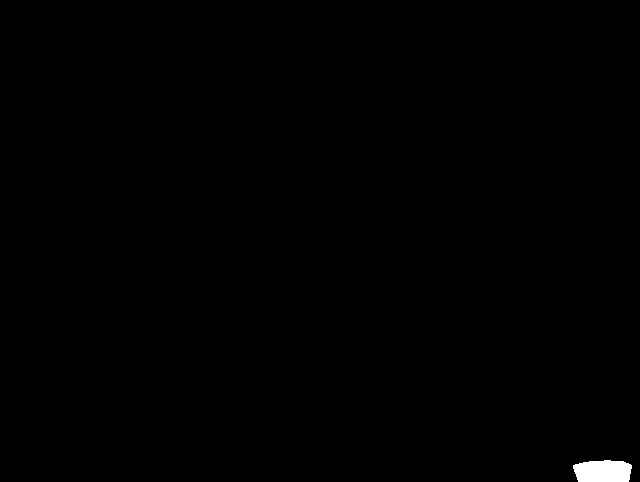}} & 0.045 \\
\includegraphics[width=0.33\textwidth]{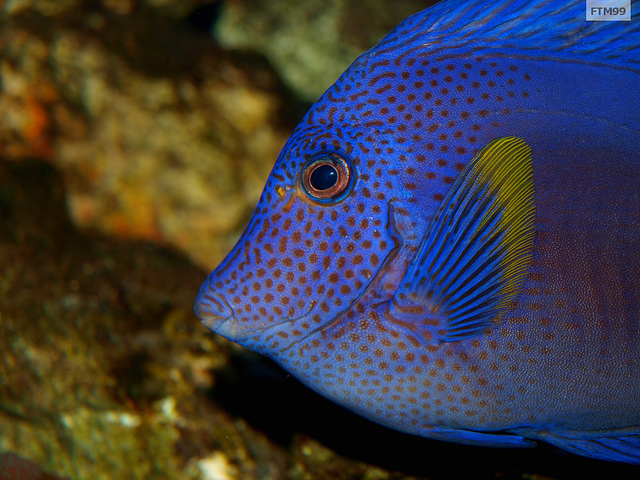} &
\fbox{\includegraphics[width=0.33\textwidth]{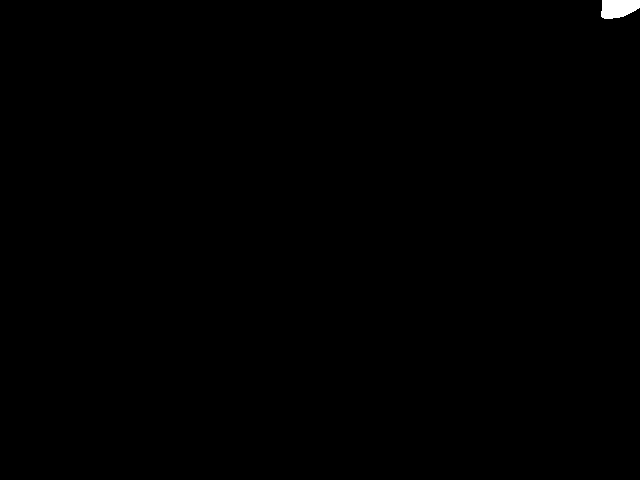}} & 0.023 \\
\hline
\end{tabular}
\end{table*}

\end{document}